\pdfoutput=1

\documentclass[11pt]{article}

\usepackage{EACL2023}

\usepackage{times}
\usepackage{latexsym}

\usepackage{inconsolata}

\usepackage[utf8]{inputenc}
\usepackage{adjustbox}
\usepackage{amsmath}
\usepackage{amssymb}
\usepackage{graphicx}
\usepackage{tabularx}
\usepackage{algorithm}
\usepackage{multirow}
\usepackage{tabularx}
\usepackage{color, colortbl}
\usepackage{caption}
\usepackage{booktabs}
\usepackage{xspace}
\usepackage{arydshln}
\usepackage{subcaption}
\usepackage{mdwlist}
\usepackage{mwe}
\usepackage{soul}
\usepackage{url}
\usepackage{amsfonts}
\usepackage{bm}
\usepackage{euflag}
\usepackage[T1]{fontenc}
\usepackage[utf8]{inputenc}
\definecolor{Gray}{gray}{0.92}
\definecolor{LightGray}{gray}{0.96}
\definecolor{LightCyan}{rgb}{0.92,0.968,0.968}
\newcolumntype{Y}{>{\centering\arraybackslash}X}
\usepackage{microtype}

\newcommand{\mbert}{{\textsc{mBERT}}\xspace}
\newcommand{\xlmr}{{\textsc{XLM-R}}\xspace}
\newcommand{\mpnet}{{\textsc{xMPNET}}\xspace}
\newcommand{\labse}{{\textsc{LaBSE}}\xspace}
\newcommand{\vecmap}{{\textsc{VecMap}}\xspace}
\newcommand{\mneg}{{\textsc{mneg}}\xspace}

\newcommand{\en}{{\textsc{en}}\xspace}
\newcommand{\de}{{\textsc{de}}\xspace}
\newcommand{\et}{{\textsc{et}}\xspace}
\newcommand{\ka}{{\textsc{ka}}\xspace}
\newcommand{\hr}{{\textsc{hr}}\xspace}
\newcommand{\he}{{\textsc{he}}\xspace}
\newcommand{\ca}{{\textsc{ca}}\xspace}
\newcommand{\ita}{{\textsc{it}}\xspace}
\newcommand{\fin}{{\textsc{fi}}\xspace}
\newcommand{\ru}{{\textsc{ru}}\xspace}
\newcommand{\tr}{{\textsc{tr}}\xspace}
\newcommand{\blg}{{\textsc{bg}}\xspace}
\newcommand{\fr}{{\textsc{fr}}\xspace}

\newcommand{\rparagraph}[1]{\vspace{1.6mm}\noindent\textbf{#1.}}
\newcommand{\sparagraph}[1]{\vspace{0.0mm}\noindent\textbf{#1.}}

\usepackage{pifont} 
\newcommand{\one}{\ding{172}\hspace{0.3mm}}
\newcommand{\two}{\ding{173}\hspace{0.3mm}}
\newcommand{\three}{\ding{174}\hspace{0.3mm}}
\newcommand{\four}{\ding{175}\hspace{0.3mm}}
\newcommand{\five}{\ding{176}\hspace{0.3mm}}
\newcommand{\six}{\ding{177}\hspace{0.3mm}}
\newcommand{\seven}{\ding{178}\hspace{0.2mm}}

\usepackage{todonotes}
\makeatletter
\newcommand*\iftodonotes{\if@todonotes@disabled\expandafter\@secondoftwo\else\expandafter\@firstoftwo\fi}
\makeatother

\definecolor{edolime}{rgb}{0.9,1,0.3}

\title{Probing Cross-Lingual Lexical Knowledge \\ from Multilingual Sentence Encoders}

\author{Ivan Vuli\'{c}$^1$ ~~~ Goran Glava\v{s}$^{2}$ ~~~ Fangyu Liu$^{1}$ \\ {\bf ~~ Nigel Collier$^1$ ~~~ Edoardo Maria Ponti$^{3,1}$ ~~~ Anna Korhonen$^{1}$}\smallskip \\
$^1$Language Technology Lab, TAL, University of Cambridge \\
$^2$CAIDAS, University of Würzburg \\
$^3$EdinburghNLP, University of Edinburgh\\
\texttt {\{iv250, alk23\}@cam.ac.uk}
}

\begin{document}
\maketitle
\begin{abstract}
Pretrained multilingual language models (LMs) can be successfully transformed into multilingual sentence encoders (SEs; e.g., \labse, \mpnet) via additional fine-tuning or model distillation with parallel data. However, it remains unclear how to best leverage them to represent sub-sentence \textit{lexical} items (i.e., words and phrases) in cross-lingual lexical tasks. In this work, we \textit{probe} SEs for the amount of \textit{cross-lingual lexical knowledge} stored in their parameters, and compare them against the original multilingual LMs. We also devise a simple yet efficient method for \textit{exposing} the cross-lingual lexical knowledge by means of additional fine-tuning through inexpensive contrastive learning that requires only a small amount of word translation pairs. Using bilingual lexical induction (BLI), cross-lingual lexical semantic similarity, and cross-lingual entity linking as lexical probing tasks, we report substantial gains on standard benchmarks (e.g., +10 Precision@1 points in BLI). The results indicate that the SEs such as \labse can be `rewired' into effective cross-lingual lexical encoders via the contrastive learning procedure, and that they contain more cross-lingual lexical knowledge than what `meets the eye' when they are used as off-the-shelf SEs. This way, we also provide an effective tool for harnessing `covert' multilingual lexical knowledge hidden in multilingual sentence encoders.
\end{abstract}


\section{Introduction}
\label{s:intro}
Transfer learning with pretrained Language Models (LMs) such as BERT \citep{devlin2019bert} and RoBERTa \citep{liu:2019roberta} offers unmatched performance in many NLP tasks \cite{Wang:2019superglue,Raffel:2019:arxiv}. However, despite the wealth of semantic knowledge stored in the pretrained LMs \cite{Rogers:2020arxiv,Vulic:2020emnlp}, they do not produce coherent and effective sentence representations when used off-the-shelf \cite{Liu:2021emnlp}: to this effect, further specialization for sentence-level semantics -- not unlike the standard task fine-tuning -- is needed \cite[\textit{inter alia}]{Reimers:2019emnlp,Li:2020emnlp,Yan:2021acl}. 
LMs get \textit{transformed} into sentence encoders (SEs) via dual-encoder frameworks that leverage contrastive learning objectives \cite{infonce,Musgrave:2020eccv}, in supervised (i.e., leveraging labeled external data such as NLI or sentence similarity annotations) \cite{Reimers:2019emnlp,Vulic:2021emnlp,Liu:2021dialoguecse} or, more recently, fully unsupervised fine-tuning \cite{Liu:2021emnlp,Gao:2021emnlp} setups.

Following the procedures from monolingual setups, another line of research has been transforming multilingual LMs into \textit{multilingual SEs} \cite{Feng:2020labse,Reimers:2020emnlp}, which enable effective sentence matching and ranking in multiple languages as well as cross-lingually \cite{Litschko:2022jir}. The transformation is typically done by coupling \textbf{1)} LM objectives on monolingual data available in multiple languages with \textbf{2)} cross-lingual objectives such as Translation Language Modeling (TLM) \cite{Conneau:2019neurips} and/or cross-lingual contrastive ranking \cite{Yang:2020demos}. Such multilingual SEs consume a large number of parallel sentences for the latter objectives. Consequently, they outperform multilingual off-the-shelf LMs in cross-lingual sentence similarity and ranking applications \cite{Liu:2021acl,Litschko:2022jir}. However, as we show in this work, such multilingual SEs may still lag behind traditional \textit{static} cross-lingual word embeddings (CLWEs) when encoding sub-sentence \textit{lexical items} (e.g., words or phrases) \cite{Liu:2021emnlp} for cross-lingual lexical tasks (e.g., BLI).

\begin{figure}[!t]
    \centering
    \includegraphics[width=0.85\linewidth]{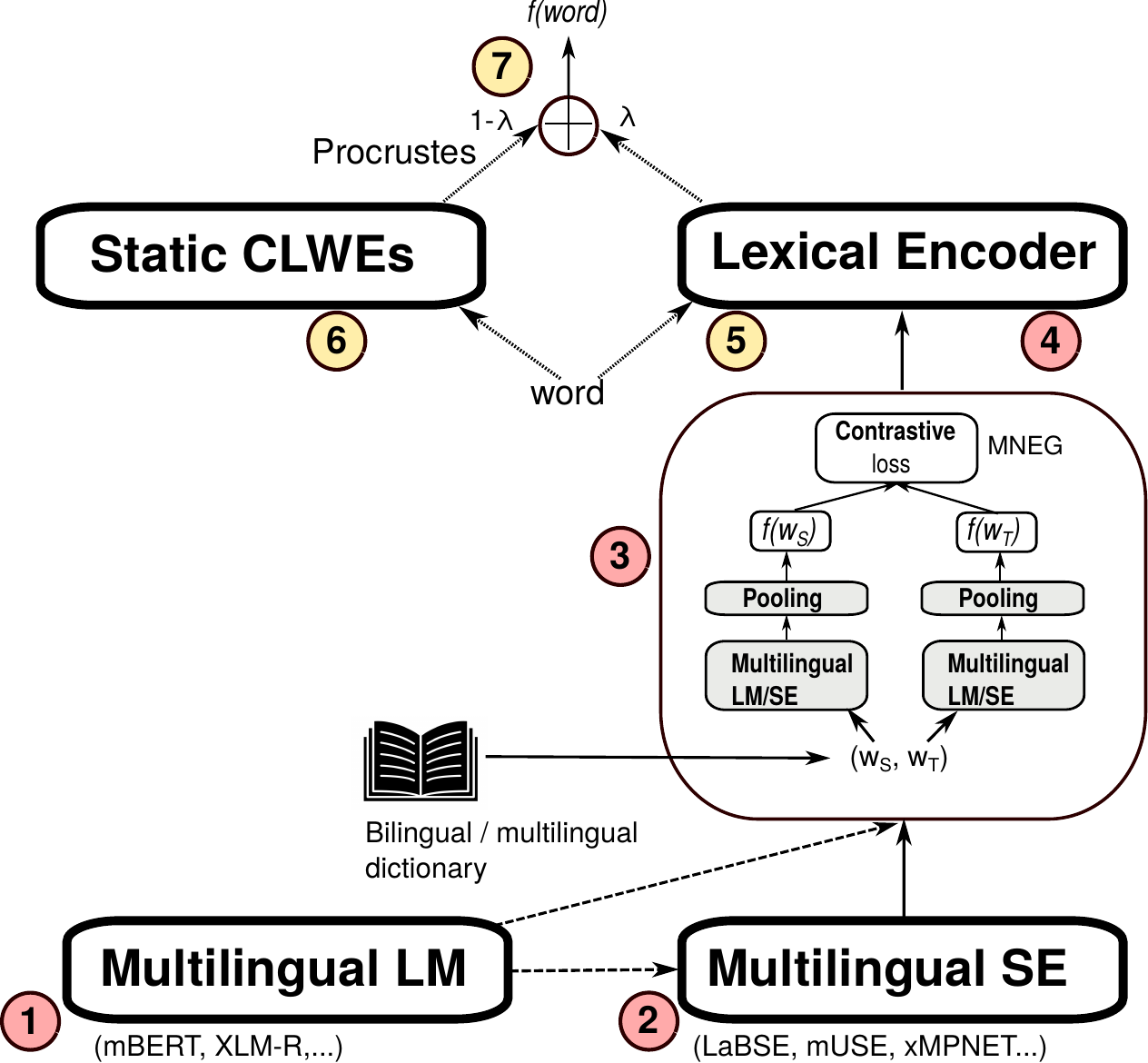}
    \caption{Illustration of the pipeline of exposing cross-lingual lexical knowledge from multilingual language models (LMs) and sentence encoders (SEs) (\S\ref{s:methodology}). Multilingual LMs (\one) can be transformed into multilingual SEs (\two) as done in previous work \cite{Reimers:2020emnlp,Feng:2020labse}. A contrastive cross-lingual lexical fine-tuning procedure (\three) (requiring an external bilingual dictionary) can be applied on both \one and \two, yielding a fine-tuned cross-lingual lexical encoder (\four). 
    At inference, a word/phrase is encoded by the lexical encoder (\five). In addition, its encoding can be interpolated with the corresponding static (cross-lingual) word embedding (\six), producing the final embedding of the word/phrase (\seven). Before the interpolation, static CLWEs must be mapped into the vector space of the lexical encoder (\four): to this end, we learn the  standard orthogonal (Procrustes) projection matrix. 
    }
    \label{fig:front_pic}
    \vspace{-2.5mm}
\end{figure}

In this work, we \textit{probe} multilingual SEs for \textit{cross-lingual lexical knowledge}. We demonstrate that, due to their fine-tuning on multilingual and parallel data, they indeed store a wealth of such knowledge, much more than what `meets the eye' when they are used `off the shelf'. 
However, this lexical knowledge needs to be \textit{exposed} from the original multilingual SEs, (again) through additional fine-tuning. In other words, we show that multilingual SEs can be `rewired' into effective cross-lingual \textit{lexical encoders}, as illustrated in Figure~\ref{fig:front_pic}. This rewiring is again done via a quick and inexpensive contrastive learning procedure: with merely 1k-5k word translation pairs, we successfully convert multilingual SEs into state-of-the-art bilingual lexical encoders for any language pair.\footnote{Note that this is a typical requirement of standard mapping-based approaches for learning static cross-lingual word embeddings which excel in the BLI task \cite{Mikolov:2013arxiv,Conneau:2018iclr,glavavs2020non}.}

We probe the original LMs and SEs as well as demonstrate the usefulness of the proposed contrastive procedure for `exposing' cross-lingual lexical knowledge on three standard lexical cross-lingual tasks using standard evaluation data and protocols: BLI, cross-lingual lexical semantic similarity (XLSIM), and cross-lingual entity linking (XL-EL). We show that the `exposure' procedure is highly effective for both vanilla multilingual LMs (mBERT and XLM-R) and multilingual SEs (\labse and \mpnet): e.g., we observe $\approx$+10 Precision@1 points gains on standard BLI benchmarks \cite{Glavas:2019acl}. Multilingual SEs offer substantially better cross-lingual lexical performance than vanilla LMs, both before and after being subjected to contrastive cross-lingual lexical fine-tuning (see Figure~\ref{fig:front_pic}). This indicates that multilingual SEs have more cross-lingual lexical knowledge than their vanilla LM counterparts, likely owing to their additional exposure to parallel data.      


Finally, inspired by \newcite{Li:2022acl}, we validate that word vectors produced by cross-lingual lexical encoders (i.e., after contrastive cross-lingual lexical `exposure') 
can be effectively interpolated with static CLWEs \cite{Artetxe:2018acl} and offer even stronger performance in cross-lingual lexical tasks. 
Encouragingly, our cross-lingual lexical specialization of multilingual SEs (as well as the further interpolation with static CLWEs), yields particularly massive performance gains for pairs of low-resource languages, as demonstrated on the low-resource BLI benchmark \cite{Vulic:2019we}.  




\section{From Multilingual Sentence Encoders to Cross-Lingual Lexical Encoders}
\label{s:methodology}
\label{ss:methodology}

\sparagraph{Motivation}
The motivation for this work largely stems from the research on \textit{probing and analyzing} pretrained LMs for various types of knowledge they (implicitly) store in their parameters \cite{Ethay:2019emnlp,Jawahar:2019acl,Rogers:2020arxiv}. In this paper, we focus on a \textit{particular knowledge type}: cross-lingual lexical knowledge, and its extraction from multilingual LMs and SEs.

Previous work tried to prompt multilingual LMs for word translations via masked natural language templates \cite{gonen-etal-2020-greek} and extract type-level word embeddings from LMs (i) directly without context \cite{vulic2020multi,Vulic:2021acl} or (ii) by averaging contextual embeddings over a large auxiliary corpus in the target language \cite{Bommasani:2020acl,Litschko:2022jir}. This existing body of work \textbf{1)} demonstrated that even sophisticated templates and extraction strategies cannot outperform cross-lingual word embedding spaces (e.g., induced from monolingual fastText vectors) in cross-lingual lexical tasks such as BLI \cite{Vulic:2020emnlp} and \textbf{2)} did not attempt to expose cross-lingual lexical knowledge from multilingual SEs and compare it against the (same type of) knowledge extracted from vanilla multilingual LMs.   



\rparagraph{Multilingual Sentence Encoders}
Off-the-shelf LMs contextualize (sub)word representations, but are unable to encode the precise meaning of input text out of the box. SEs -- LMs fine-tuned via sentence-level objectives -- in contrast, directly produce a precise semantic encoding of input text.
A large body of work focuses on inducing multilingual encoders that capture sentence meaning across languages \cite[\textit{inter alia}]{Artetxe:2019tacl,Feng:2020labse,Yang:2020demos}. 

The most popular approach obtains multilingual SEs \cite{Reimers:2020emnlp} by distilling the knowledge from the monolingual English SE teacher (trained on English semantic similarity and NLI data) into multilingual LM student (e.g., mBERT), using parallel sentences to guide the distillation process. SEs, being specialized for sentence similarity, encode sentence meaning more accurately and are useful in various (unsupervised) text similarity and ranking tasks, monolingually and across languages.

While SEs' primary purpose is sentence encoding, they can, in principle, be applied to sub-sentential text: words and phrases. In this work, we show that multilingual SEs can be turned into effective cross-lingual lexical encoders. We achieve this through additional cross-lingual lexical fine-tuning \cite{Vulic:2021acl}, requiring as supervision only a small set of word translation pairs.

\subsection{Cross-Lingual Lexical Fine-Tuning}
\label{ss:finetuning}

For a given language pair $L_s$-$L_t$, our contrastive cross-lingual lexical fine-tuning of multilingual encoders (LMs and SEs alike) requires a dictionary spanning $N$ (typically $N$\,$\leq$\,5,000) word translation pairs, $\mathcal{D}=\{(w_{i,s},v_{i,t})\}_{i = 1}^N$.\footnote{Note that such bilingual dictionaries are one of the most widespread and cheapest-to-obtain resources in multilingual NLP \cite{Ruder:2019jair,Wang:2022acl}.} We consider the translation pairs from $\mathcal{D}$ to be \textit{positive examples} for the contrastive fine-tuning procedure. 
For each of the $N$ source language words in the dictionary ($w_{i,s}$), we precompute a set of $N_n$ hard negative samples: these are the $L_t$ words that are the closest to $w_{i,s}$ in the representation space of the multilingual encoder, but not its direct translation $v_{i,t}$. Let $\textit{f}_{\theta}(\cdot)$, be the encoding function of the multilingual LM/SE, with $\theta$ as parameters, and let $S(\cdot, \cdot)$ be a function of similarity between two vectors. For a source word $w_{i,s}$, we select as hard negatives words $v_t$ from $L_t$ that have the highest $S(\textit{f}_{\theta_0}(w_{i,s}), \textit{f}_{\theta_0}(v_t))$ score, with $\theta_0$ as the original encoder's parameters, before fine-tuning.  

We encode all training words -- those from the seed dictionary $\mathcal{D}$ and (at most) $N\cdot N_n$ precomputed hard $L_t$ negatives -- independently and \textit{in isolation}. Concretely, for an input $w$ with $M$ subword tokens $[sw_1]\ldots[sw_M]$, we feed the sequence $[SPEC1] [sw_1]\ldots[sw_M] [SPEC2]$ into the multilingual encoder (with $[SPEC1]$ and $[SPEC2]$ as encoder's sequence start and end tokens, resp.), and take the average of the transformed representation (from the last Transformer layer) of the $w$'s subword tokens as the $w$'s encoding $\textit{f}_{\theta}(w)$.\footnote{Put simply, we process sub-sentential text input in the same way that multilingual SEs handle sentence-level input. We experimented with other encoding strategies from prior work, e.g., taking the representation of the sequence start token $[SPEC1]$ \cite{Liu:2021naacl,Li:2022acl}; in preliminary experiments, however, we obtained the best results by averaging subword representations. Note that $[SPEC1]$ and $[SPEC2]$ are placeholders for the encoder's special tokens: e.g., in case of multilingual BERT [SPEC1] is the [CLS] token, while [SPEC2] is the [SEP] token.} 


Following common practice in contrastive learning \cite{Henderson:2019acl,Vulic:2021acl}, we instantiate $S$ as the scaled cosine similarity: $S(\textit{f}_{\theta}(w_i), \textit{f}_{\theta}(w_j)) = C\cdot cos(\textit{f}_{\theta}(w_i), \textit{f}_{\theta}(w_j))$, with $C$ as the scaling constant. We then train in batches of $B$ translation pairs, with the variant of the widely used multiple negatives ranking loss (\textsc{mneg}) \cite{Cer:2018arxiv,Henderson:2019acl,Henderson:2019convert} as the fine-tuning objective:


\vspace{-3.5mm}
{\small
\begin{align}
\mathcal{L} = &-\sum_{i=1}^B S(\textit{f}_{\theta}(w_i),\textit{f}_{\theta}(v_i)) \hspace{8mm} \text{(positives)} \notag \\
&+ \sum_{i=1}^B \log \hspace{-0.7em} \sum_{j=1,j\neq i}^{B} \hspace{-0.7em}e^{S(\textit{f}_{\theta}(w_i),\textit{f}_{\theta}(v_j))} \hspace{2mm} \text{(in-batch\,negatives)} \notag \\
&+ \sum_{i=1}^B \log \sum_{k=1}^{N_n} e^{S(\textit{f}_{\theta}(w_i),\textit{f}_{\theta}(v_{k,i}))} \hspace{5mm} \text{(hard\,negatives)} \notag
\label{eq:mneg}
\end{align}
}%
where $v_{k,i}$ denotes the $k$-th hard negative from the language $L_t$ for the $L_s$ word $w_i$. \mneg combines the $N_n$ hard negatives per each positive example with $B$-1 in-batch negatives (i.e., for a source language word $w_{i, k}$, each target language word $v_{j,t}$, $j \neq i$ from $B$ is used as an in-batch negative of $w_{i, k}$). \textsc{mneg} aims to reshape the representation space of the encoder by simultaneously (a) maximising the similarity for positive pairs -- i.e., bringing closer together (`attracting') the words from the positive pairs and (b) minimising the similarity for (both in-batch and hard) negative pairs -- i.e., pushing (`repelling') the words from negative pairs further away from each other).\footnote{In practice, we rely on the implementation of the \mneg loss from the SBERT repository \url{www.sbert.net} \cite{Reimers:2019emnlp}; the default value $C=20$ is used.}

\subsection{Interpolation with Static CLWEs}
\label{ss:interpolation}

\citet{Li:2022acl} recently showed that further performance gains in the BLI task might be achieved by combining the type-level output of the encoding function $\textit{f}$ with static CLWEs, but they experimented only with multilingual LMs. Static CLWEs and multilingual encoder-based representations of the same set of words can be seen as two different views of the same data point. Following \citet{Li:2022acl}, we learn an additional linear orthogonal mapping from the static cross-lingual WE space -- e.g., a CLWE space induced from monolingual fastText embeddings \cite{Bojanowski:2017tacl} using \vecmap \cite{Artetxe:2018acl} -- into the cross-lingual space spanned by the multilingual encoder. The mapping transforms $\ell_2$-normed $d_1$-dimensional static CLWEs into $d_2$-dimensional cross-lingual WEs obtained through the multilingual encoder (fine-tuned $\textit{f}_{\theta}$ or original $\textit{f}_{\theta_0}$).

Learning the linear map $\bm{W}$$\in$$\mathbb{R}^{d_{1}\times d_{2}}$, when $d_1 < d_2$,\footnote{The assumption $d_1$ < $d_2$ typically holds as fastText WEs are 300-dimensional while the dimensionality of standard multilingual LMs and SEs is $d_2=768$ or $d_2=1,024$.} is formulated as a Generalized Procrustes problem \cite{schonemann1966generalized,viklands2006algorithms}. It operates on all (i.e., both $L_s$ and $L_t$) words from the seed translation dictionary $\mathcal{D}$.\footnote{To learn the mapping $\bm{W}$, for pairs from $\mathcal{D}$ we decouple $L_s$ words $w_{i,s}$ from their $L_t$ translations $v_{i,t}$ to create vector pairs $(\mathit{clwe}(w_{i,s}), \textit{f}_{\theta}(w_{i,s}))$ and $(\mathit{clwe}(v_{i,t}), \textit{f}_{\theta}(v_{i,t}))$ -- with $\mathit{clwe}(w)$ as the static CLWE of $w$ (e.g., its \vecmap embedding), and $\textit{f}_{\theta}(w)$ its encoder-based representation -- based on which we learn of the orthogonal mapping $\bm{W}$ (the so-called Procrustes method gives a closed-form solution).}
Unless noted otherwise, a final representation of an input word $w$ is then computed as follows:

\vspace{-2.5mm}
{\footnotesize
\begin{align}
(1-\lambda)\frac{\mathit{clwe}(w)\bm{W}}{\left\|\mathit{clwe}(w)\bm{W}\right\|_{2}} + \lambda \frac{f_{\theta}(w)}{\left\|f_{\theta}(w)\right\|_{2}},
\end{align}
}

\noindent where $\lambda$ is a tunable interpolation hyper-parameter, $\mathit{clwe}(w)$ denotes the static CLWE of $w$, and $f_{\theta}(w)$ the representation of $w$ obtained with the (contrastively fine-tuned or original) multilingual LM/SE. This simple procedure yields an `interpolated' shared cross-lingual WE space.

\section{Experimental Setup}
\label{s:exp}
\sparagraph{Multilingual Sentence Encoders} We probe two widely used multilingual SEs:  \textbf{1)} Language-agnostic BERT Sentence Embedding (\textbf{\labse}) \cite{Feng:2020labse} which adapts pretrained multilingual BERT (\textbf{\mbert}) \cite{devlin2019bert} into a multilingual SE; \textbf{2)} Multilingual \textbf{\mpnet} is a distillation-based adaptation \cite{Reimers:2020emnlp} of \textbf{\xlmr} \cite{Conneau:2020acl} as the student model into a multilingual SE, based on the monolingual English MPNet encoder \cite{Song:2020neurips} as the teacher model. \labse is the current state-of-the-art multilingual SE and supports 109 languages, while \mpnet is the best-performing multilingual SE in the Sentence-BERT repository \cite{Reimers:2019emnlp}.\footnote{For further technical details regarding the models in our comparison, we refer the reader to the original papers.} Along with \labse and \mpnet as SEs, we experiment with the original multilingual LMs -- mBERT and XLM-R -- using the the same training and evaluation protocols (see Figure~\ref{fig:front_pic} and \S\ref{s:methodology}), aiming to quantify: (i) the extent to which cross-lingual lexical knowledge can be exposed from LMs that have \textit{not} been specialized for sentence-level semantics, as well as (ii) the increase in quality of lexical knowledge brought about with sentence-level specialisation (i.e., when multilingual LMs get transformed into multilingual SEs).


 
\rparagraph{Evaluation Tasks}
We evaluate on the standard and diverse cross-lingual lexical semantic tasks:

\vspace{1.2mm}
\noindent \textbf{Task 1: Bilingual Lexicon Induction (BLI)}, a standard task to assess the ``semantic quality'' of static cross-lingual word embeddings (CLWEs) \cite{Ruder:2019jair}, allows us to \textbf{1)} directly assess the extent to which cross-lingual word translation knowledge can be exposed from multilingual LMs and SEs and \textbf{2)} immediately test the ability to transform multilingual sentence encoders into bilingual lexical encoders; 
We run a series of BLI evaluations on two standard BLI evaluation benchmarks. First, GT-BLI \cite{Glavas:2019acl}, constructed semi-automatically from Google Translate, comprises 28 language pairs with a good balance of typologically similar and distant languages (Croatian: \textsc{hr}, English: \textsc{en}, Finnish: \textsc{fi}, French: \textsc{fr}, German: \textsc{de}, Italian: \textsc{it}, Russian: \textsc{ru}, Turkish: \textsc{tr}). Second, PanLex-BLI \cite{Vulic:2019we} focuses on BLI evaluation for lower-resource languages, deriving training and test data from PanLex \cite{Kamholz:2014lrec}. We evaluate on 10 pairs comprising the following five typologically and etymologically diverse languages: Bulgarian (\blg), Catalan (\ca), Estonian (\et), Hebrew (\he), and Georgian (\ka). 

Standard BLI setups and data are adopted: 5k training word pairs are used as seed dictionary $\mathcal{D}$, and another 2k pairs as test data. Note that $\mathcal{D}$ is used to (i) contrastively fine-tune multilingual encoders (\S\ref{ss:finetuning}), (ii) learn the (baseline) static \vecmap CLWE space, as well as to (iii) learn the projection between the static CLWE space and the representation spaces of multilingual encoders required to obtain the interpolated representations (\S\ref{ss:interpolation}). The evaluation metric is standard Precision@1 (P@1).\footnote{We observed very similar performance trends for P@5 and Mean Reciprocal Rank (MRR) as BLI measures.} For PanLex-BLI, we also run additional experiments using smaller bilingual dictionaries $\mathcal{D}$, spanning 1k translation pairs.

\vspace{1.2mm}
\noindent \textbf{Task 2: Cross-Lingual Lexical Semantic Similarity (XLSIM)} is another cross-lingual lexical task: it tests the extent to which lexical representations can capture the (human perception of) fine-grained semantic similarity of words across languages. We use the recent comprehensive XLSIM benchmark Multi-SimLex \cite{vulic2020multi}, which comprises cross-lingual datasets of 2k-4k scored word pairs over 66 language pairs. We evaluate on a subset of language pairs shared with the GT-BLI dataset: \en, \fin, \ru, \fr.\footnote{The evaluation metric is the standard Spearman's rank correlation between the average of gold human-elicited XLSIM scores for word pairs and the cosine similarity between their respective word representations. To avoid any test data leakage, we remove all XLSIM test pairs from the bilingual dictionary $\mathcal{D}$ prior to fine-tuning and CLWE mapping.}

\vspace{1.2mm}
\noindent \textbf{Task 3: Cross-Lingual Entity Linking (XEL)} is a standard task in knowledge base (KB) construction \cite{Zhou:2022acl}, where the goal is to link an entity mention in any language to a corresponding entity in an English KB or in a language-agnostic KB.\footnote{Following prior work \cite{Liu:2021naacl,Zhou:2022acl}, XEL in this work also refers only to entity mention \textit{disambiguation}; it does not cover the mention detection subtask.} We evaluate on the cross-lingual biomedical entity linking (XL-BEL) benchmark of \citet{Liu:2021acl}: it requires the model to link an entity mention to entries in UMLS \citep{bodenreider2004unified}, a language-agnostic medical knowledge base. We largely follow the XL-BEL experimental setup of \citet{Liu:2021acl} and probe the encoders first \textit{without} any additional task-specific fine-tuning on UMLS data, and then \textit{with} subsequent UMLS fine-tuning (i) only on the \en UMLS data, (ii) on all the UMLS data in 10 languages of the XL-BEL dataset.\footnote{See \cite{Liu:2021acl} for additional details.} Due to a large number of experiments, we again focus on the subset of languages in XL-BEL shared with GT-BLI: \en, \de, \fin, \ru, \tr. 


\begin{table*}[t]
\def\arraystretch{0.9}
\centering
{\scriptsize
\begin{tabularx}{\linewidth}{l Y YYYY YYYY}
\toprule
{\bf Multilingual LMs} & {} & \multicolumn{4}{c}{\bf \mbert} & \multicolumn{4}{c}{\bf \xlmr} \\
\cmidrule(lr){3-6} \cmidrule(lr){7-10}
\rowcolor{Gray}
{\bf Config} $\xrightarrow{}$ & {\bf \vecmap} & {noCL (1.0)} & {noCL ($\lambda$)} & {+CL (1.0)} & {+CL ($\lambda$)} & {noCL (1.0)} & {noCL ($\lambda$)} & {+CL (1.0)} & {+CL ($\lambda$)} \\
\cmidrule(lr){2-2} \cmidrule(lr){3-6} \cmidrule(lr){7-10}
{\textbf{[BLI]} $\lambda$=0.3} & {42.7} & {9.0} & {39.2} & {22.3} & {44.3} & {6.4} & {33.7} & {21.2}  & {43.8} \\
\hdashline
{\textbf{[XLSIM]} $\lambda$=0.5} & {45.8} & {5.7} & {35.4} & {38.4} & {48.1} & {1.7} & {23.5} & {46.1}  & {51.8} \\
\midrule
\midrule
{\bf Multilingual SEs} & {} & \multicolumn{4}{c}{\bf \labse} & \multicolumn{4}{c}{\bf \mpnet} \\
\cmidrule(lr){3-6} \cmidrule(lr){7-10}
\rowcolor{Gray}
{\bf Config} $\xrightarrow{}$ & {\bf \vecmap} & {noCL (1.0)} & {noCL ($\lambda$)} & {+CL (1.0)} & {+CL ($\lambda$)} & {noCL (1.0)} & {noCL ($\lambda$)} & {+CL (1.0)} & {+CL ($\lambda$)} \\
\cmidrule(lr){2-2} \cmidrule(lr){3-6} \cmidrule(lr){7-10}
{\textbf{[BLI]} $\lambda$=0.3} & {42.7} & {21.4} & {45.7} & {30.8} & {49.1} & {17.0} & {41.7} & {28.6}  & {47.9} \\
\hdashline
{\textbf{[XLSIM]} $\lambda$=0.5} & {45.8} & {50.4} & {54.9} & {48.8} & {54.1} & {51.3} & {56.6} & {49.6}  & {54.5} \\
\bottomrule
\end{tabularx}
}%
\vspace{-1mm}
\caption{(a) P@1 scores ($\times$100\%) averaged across all 28 language pairs in the GT-BLI dataset ([BLI] rows); (b) Spearman's $\rho$ correlaction scores ($\times$100) averaged across a subset of 6 language pairs from Multi-SimLex ([XLSIM rows]). See \S\ref{s:exp} for the description of different model configurations/variants. $|\mathcal{D}|=5k$. The number in the parentheses denotes the value for $\lambda$ (see \S\ref{s:exp}), which differs between the two tasks (0.3 for BLI and 0.5 for XLSIM). The $\lambda$ value of 1.0 effectively means 'no interpolation' with static \vecmap CLWEs. Individual results per each language pair in both tasks and with other $\lambda$s are in Appendix~\ref{app:gtbli} and Appendix~\ref{app:msimlex}.}
\label{tab:main-gtbli}
\vspace{-1mm}
\end{table*}

\begin{table*}[t]
\def\arraystretch{0.69}
\centering
{\scriptsize
\begin{tabularx}{\linewidth}{l Y YY YY YY YY}
\toprule
{} & {} & \multicolumn{2}{c}{\bf \mbert} & \multicolumn{2}{c}{\bf \xlmr} & \multicolumn{2}{c}{\bf \labse} & \multicolumn{2}{c}{\bf \mpnet} \\
\cmidrule(lr){3-4} \cmidrule(lr){5-6} \cmidrule(lr){7-8} \cmidrule(lr){9-10}
\rowcolor{Gray}
{\bf Pair} $\downarrow$ {\bf / Config} $\xrightarrow{}$ & {\bf \vecmap} & {+CL (1.0)} & {+CL (0.4)} & {+CL (1.0)} & {+CL (0.4)} & {+CL (1.0)} & {+CL (0.4)} & {+CL (1.0)} & {+CL (0.4)} \\
\cmidrule(lr){3-4} \cmidrule(lr){5-6} \cmidrule(lr){7-8} \cmidrule(lr){9-10}
{\blg--\ca} & {34.4} & {9.6} & {31.9} & {13.2} & {33.3} & {17.9} & {\bf 38.0} & {15.9}  & {35.7} \\
{\blg--\et} & {30.0} & {17.1} & {32.6} & {21.3} & {34.1} & {29.9} & {\bf 42.7} & {26.1}  & {38.9} \\
{\blg--\he} & {26.1} & {9.9} & {21.1} & {10.5} & {26.3} & {23.7} & {\bf 37.2} & {10.9}  & {27.2} \\
{\blg--\ka} & {26.8} & {16.0} & {29.8} & {15.9} & {30.5} & {27.2} & {\bf 37.4} & {18.7}  & {32.4} \\
{\ca--\et} & {26.3} & {26.8} & {32.9} & {23.5} & {34.1} & {28.8} & {\bf 38.6} & {29.0}  & {38.9} \\
{\ca--\he} & {23.3} & {2.3} & {12.5} & {4.9} & {18.5} & {12.7} & {\bf 28.9} & {8.5}  & {22.7} \\
{\ca--\ka} & {20.7} & {1.5} & {10.3} & {4.7} & {20.0} & {9.6} & {\bf 26.1} & {6.4}  & {21.8} \\
{\et--\he} & {18.6} & {15.0} & {21.9} & {17.7} & {26.0} & {31.0} & {\bf 37.8} & {18.5}  & {27.0} \\
{\et--\ka} & {16.5} & {7.2} & {18.2} & {12.7} & {24.3} & {19.3} & {\bf 30.3} & {12.5}  & {25.8} \\
{\he--\ka} & {12.7} & {15.6} & {23.8} & {13.3} & {23.1} & {25.3} & {\bf 30.2} & {15.1}  & {24.4} \\
\cmidrule(lr){2-10}
{\bf Average} & {23.5} & {12.1} & {23.5} & {13.8} & {27.0} & {22.5} & {\bf 34.7} & {16.2}  & {29.5} \\
\bottomrule
\end{tabularx}
}%
\vspace{-1.5mm}
\caption{P@1 scores over a representative subset of 10 language pairs from the PanLex-BLI dataset of \newcite{Vulic:2019we}. See \S\ref{s:exp} for the description of different model configurations/variants. $|\mathcal{D}|=5k$. Highest scores per row are in \textbf{bold}. Respective average scores for the \textit{noCL (1.0)} config (i.e., without contrastive learning and without interpolation with static \vecmap CLWEs) are: 4.2 (\mbert), 3.1 (\xlmr), 17.0 (\labse), 8.3 (\mpnet).}
\label{tab:main-panlex}
\vspace{-1.5mm}
\end{table*}

\rparagraph{Static CLWEs and Word Vocabularies}
As monolingual static WEs, we select CommonCrawl fastText vectors \cite{Bojanowski:2017tacl} of the top 200k most frequent word types in the training data, following prior work on learning static CLWEs \cite{Conneau:2018iclr,Artetxe:2018acl,Heyman:2019naacl}.\footnote{CommonCrawl-based fastText WEs typically outperform other popular choice for monolingual WEs: Wikipedia-based fastText \cite{Glavas:2019acl,Li:2022acl}. We note that the main trends in our results also extend to the Wiki-based WEs.} Static CLWEs are then induced via the standard and popular supervised mapping-based \vecmap method \cite{Artetxe:2018acl}, leveraging the seed dictionary $\mathcal{D}$. These CLWEs are used for interpolation with encoder-based WEs (see \S\ref{ss:interpolation}) but also as the baseline approaches for BLI and XLSIM tasks.
We compute the type-level WEs from multilingual LMs and SEs for the same 200K most frequent words of each language.

\rparagraph{Technical Details and Hyperparameters} The implementation is based on the SBERT framework \cite{Reimers:2019emnlp}, using the suggested settings: AdamW \cite{Loschilov:2018iclr}; learning rate of $2e$-$5$; weight decay rate of $0.01$. We run contrastive fine-tuning for 5 epochs with all the models, with the batch size of $B=128$ positives for \mneg. The number of hard negatives per each positive is set to $N_n=10$ (see \S\ref{ss:finetuning}).\footnote{We also tested $N_n$=$\{20,30,50\}$. They slow down fine-tuning while yielding small-to-negligible performance gains.} Since standard BLI and XLSIM datasets lack a validation portion \cite{Ruder:2019jair}, we follow prior work \cite{Glavas:2019acl} and tune hyperparameters on a \textit{single, randomly selected} language pair from each dataset, and use those values in all other runs. 

\rparagraph{Model Configurations} They are labelled as follows: \textbf{ENC-\{noCL,+CL\} ($\lambda$)}, where (i) ENC denotes the input multilingual Transformer, which can be a multilingual LM (\mbert, \xlmr), or a multilingual SE (\labse, \mpnet), (ii) `noCL' refers to using the input model `off-the-shelf' without any contrastive lexical fine-tuning, while `+CL' variants apply the contrastive fine-tuning, and (iii) $\lambda$ is the factor that defines the interpolation with the static CLWE space, obtained with \vecmap (see Figure~\ref{fig:front_pic} and \S\ref{ss:interpolation}). Note that $\lambda = 1.0$ implies no interpolation with static CLWE space, i.e., WEs come purely from the multilingual LM/SE.

\rparagraph{Important Disclaimer}
We note that the main purpose of the chosen evaluation tasks and experimental protocols is not necessarily achieving state-of-the-art performance, but rather probing different model variants in different cross-lingual lexical tasks, and offering fair and insightful comparisons 


\section{Results and Discussion}
\label{s:results}

\begin{table}[!t]
\def\arraystretch{0.7}
\centering
{\scriptsize
\begin{tabularx}{\columnwidth}{l Y cYY}
\toprule
{} & {} & \multicolumn{3}{c}{\bf \labse} \\
\cmidrule(lr){3-5}
\rowcolor{Gray}
{\bf Pair} $\downarrow$ & {\bf \vecmap} & {noCL (1.0)} & {+CL (1.0)} & {+CL (0.5)} \\
\cmidrule(lr){3-5} 
{\blg--\ca} & {15.2} & {14.0} & {18.0} & {\bf 28.9} \\
{\blg--\et} & {12.5} & {20.3} & {25.5} & {\bf 35.1} \\
{\blg--\he} & {5.6} & {18.3} & {20.8} & {\bf 24.7} \\
{\blg--\ka} & {9.1} & {16.0} & {21.6} & {\bf 29.7} \\
{\ca--\et} & {9.8} & {24.8} & {25.6} & {\bf 31.2} \\
{\ca--\he} & {5.0} & {10.6} & {12.2} & {\bf 15.6} \\
{\ca--\ka} & {5.5} & {5.7} & {8.3} & {\bf 14.8} \\
{\et--\he} & {3.1} & {\bf 27.7} & {25.1} & {25.4} \\
{\et--\ka} & {4.6} & {13.2} & {16.0} & {\bf 21.1} \\
{\he--\ka} & {3.2} & {19.0} & {22.0} & {\bf 25.4} \\
\cmidrule(lr){3-5}
{\bf Average} & {7.4} & {17.0} & {19.5} & {\bf 25.2} \\
\bottomrule
\end{tabularx}
}%
\vspace{-1.5mm}
\caption{P@1 scores over 10 language pairs from the PanLex-BLI dataset of \newcite{Vulic:2019we} when $|\mathcal{D}|=1k$, with different model variants based on \labse (see \S\ref{s:exp}). Highest scores per row are in \textbf{bold}.}
\label{tab:1k-panlex}
\vspace{-2mm}
\end{table}

\noindent \textbf{Bilingual Lexicon Induction (BLI).}
Table~\ref{tab:main-gtbli} displays our main BLI results, aggregated over all 28 language pairs of GT-BLI, for two multilingual LMs (\mbert and \xlmr) and their SE counterparts (\labse and \mpnet). 
%
Two trends hold across the board. First, multilingual SEs, \labse and \mpnet, largely outperform their multilingual LM counterparts, \mbert and \xlmr. The gains are visible in all four experimental configurations (with/without contrastive cross-lingual lexical specialisation $\times$ with/without interpolation with the \vecmap
CLWE space). This confirms our intuition that multilingual SEs, having been (additionally) trained on parallel data \cite{Feng:2020labse,Reimers:2020emnlp}, should better reflect the cross-lingual alignments at the lexical level than off-the-shelf multilingual LMs, which have not been exposed to any cross-lingual signal in pretraining. The poor cross-lingual lexical alignment in the representation spaces of \mbert and \xlmr also reflects in the fact that with those encoders, we only surpass the baseline \vecmap performance by a small margin (+1.1 for \xlmr, +1.6 for \mbert) after subjecting them to contrastive lexical fine-tuning \textit{and} interpolating their word encodings with \vecmap WEs. 

The behavior of SEs, on the other hand, is much more favorable. \labse, for example, surpasses baseline \vecmap performance with interpolation alone, even without the contrastive lexical fine-tuning. When contrastively fine-tuned (and then interpolated with \vecmap) both \labse and \mpnet surpass the baseline \vecmap performance by a much wider margin (+6.4 and +5.2, respectively). This implies that our contrastive fine-tuning exposes more of the high quality cross-lingual lexical knowledge from multilingual SEs. 

Both (i) contrastive cross-lingual lexical learning (+CL) and (ii) interpolation with \vecmap consistently improve the performance for all four encoders: we reach peak scores by combining contrastive fine-tuning and interpolation with static CLWEs (+CL ($\lambda$)). Contrastive fine-tuning crucially contributes to the overall performance: compared to interpolation alone (noCL ($\lambda$)), +CL ($\lambda$) brings an average gain of over 6 BLI points.   


\captionsetup[subfigure]{oneside,margin={-0.6cm,0.8cm},skip=-5pt}
\begin{figure*}[!t]
    \centering
    \begin{subfigure}[!ht]{0.309\linewidth}
        \centering
        \includegraphics[width=0.99\linewidth]{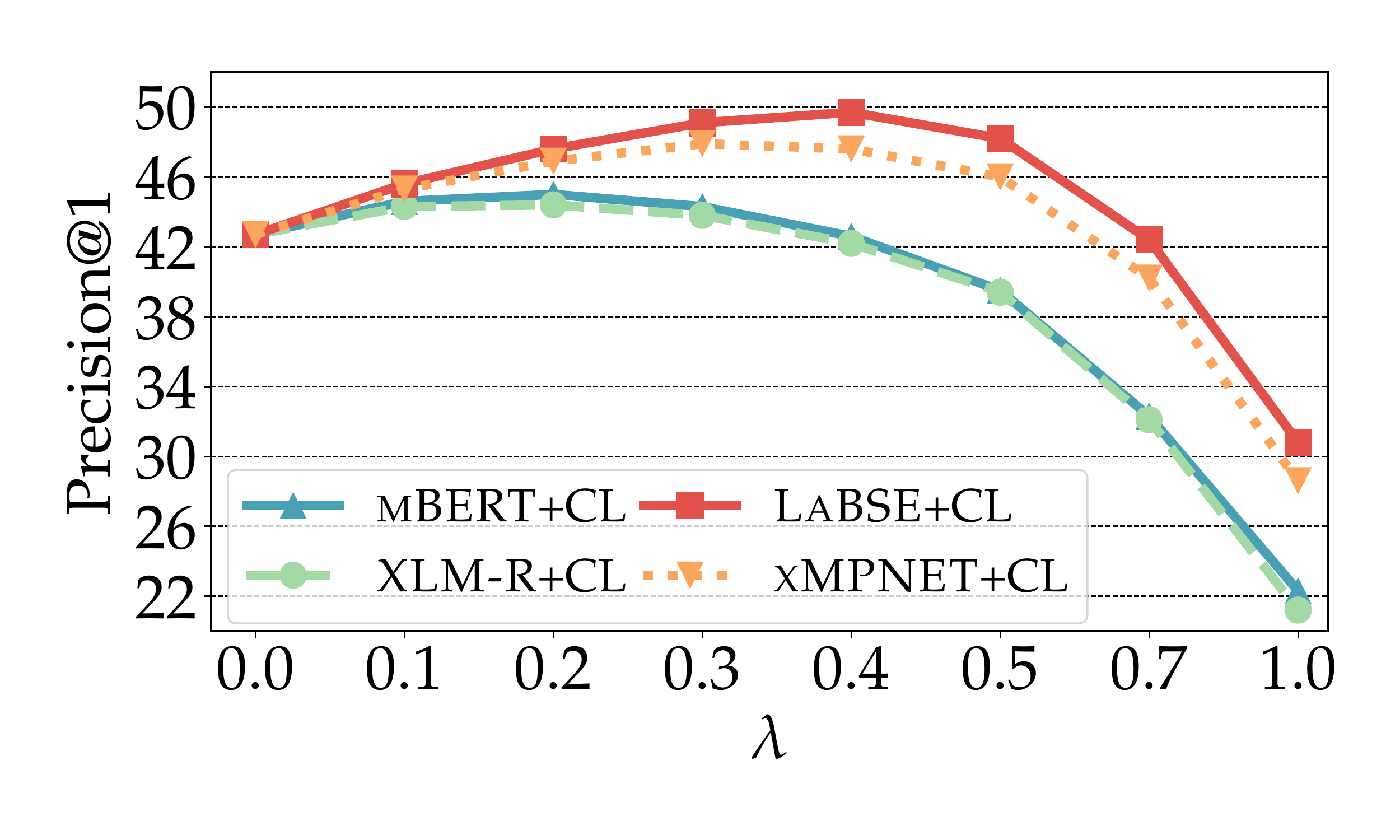}
        \caption{GT-BLI}
        \label{fig:lambda-gtbli}
    \end{subfigure}
        \begin{subfigure}[!ht]{0.309\linewidth}
        \centering
        \includegraphics[width=0.99\linewidth]{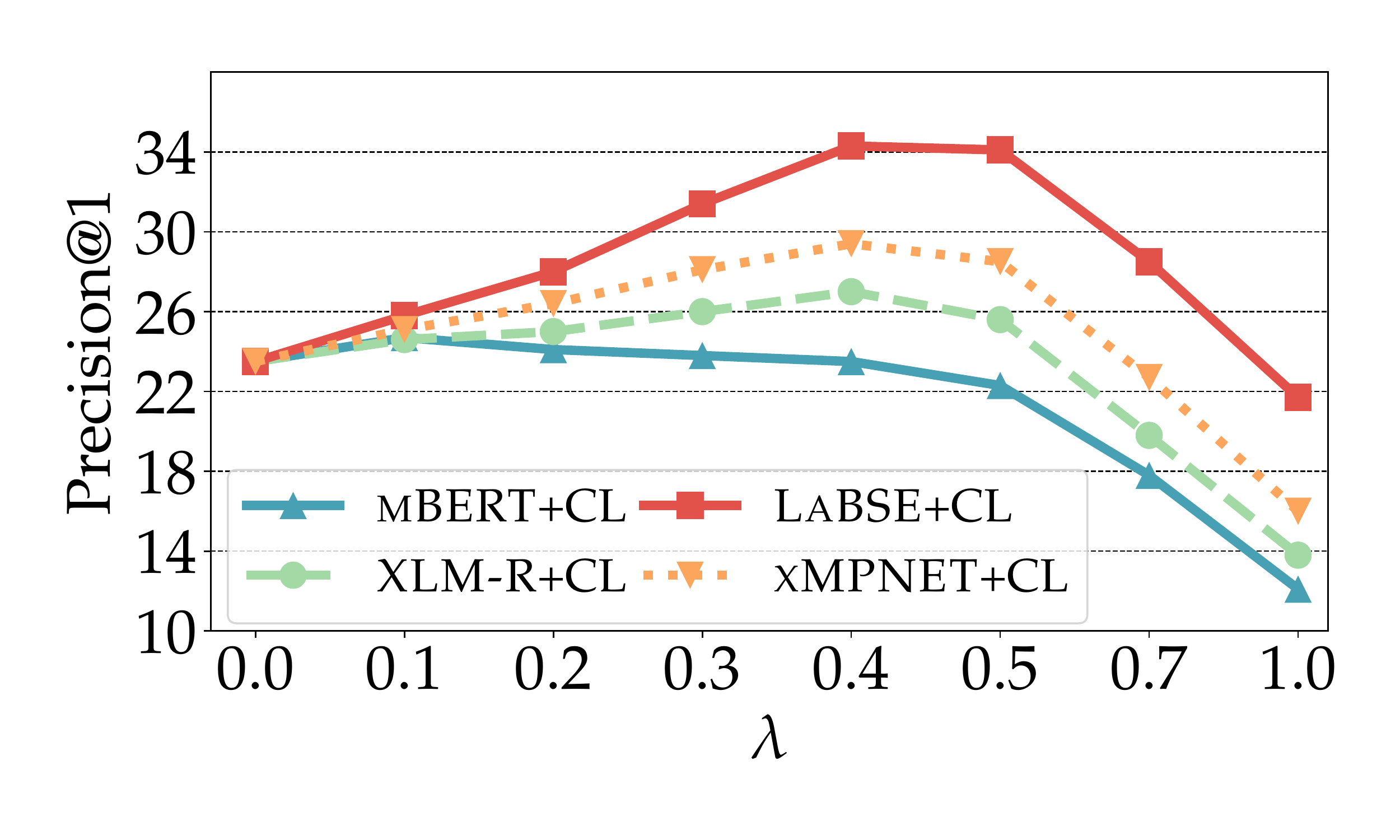}
        \caption{PanLex-BLI}
        \label{fig:lambda-panlex}
    \end{subfigure}
    \begin{subfigure}[!ht]{0.309\linewidth}
        \centering
        \includegraphics[width=0.99\linewidth]{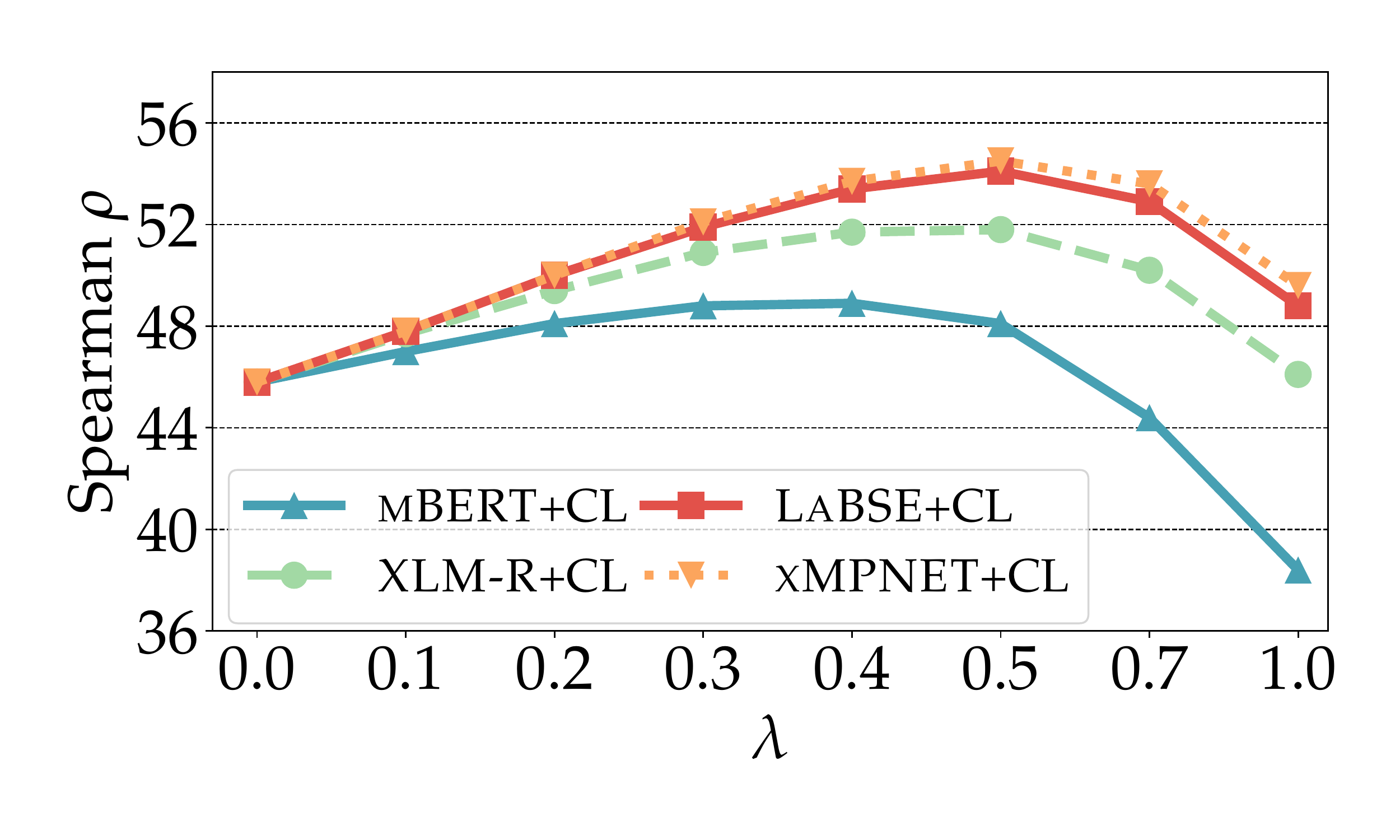}
        \caption{Multi-SimLex (XLSIM)}
        \label{fig:lambda-xlsim}
    \end{subfigure}
    \vspace{-1mm}
    \caption{Average scores across different interpolation values $\lambda$ for the BLI task on \textbf{(a)} GT-BLI and \textbf{(b)} PanLex-BLI, and \textbf{(c)} the XLSIM task on Multi-SimLex. $|\mathcal{D}|=5k$. Additional results are in Appendix~\ref{app:gtbli} and~\ref{app:msimlex}.}
    \vspace{-2mm}
\label{fig:lambdas}
\end{figure*}
\begin{table*}[t]
\def\arraystretch{0.73}
\centering
{\scriptsize
\begin{tabularx}{\linewidth}{l YY YY YY YY}
\toprule
\rowcolor{Gray}
{\bf Config} $\downarrow$ {\bf / Language $L_t$} $\xrightarrow{}$ & \multicolumn{2}{c}{\bf \de} & \multicolumn{2}{c}{\bf \fin} & \multicolumn{2}{c}{\bf \ru} & \multicolumn{2}{c}{\bf \tr} \\
\cmidrule(lr){2-3} \cmidrule(lr){4-5} \cmidrule(lr){6-7}  \cmidrule(lr){8-9} 
{} & {P@1} & {P@5} & {P@1} & {P@5} & {P@1} & {P@5} & {P@1} & {P@5} \\
\cmidrule(lr){2-3} \cmidrule(lr){4-5} \cmidrule(lr){6-7}  \cmidrule(lr){8-9} 
{\xlmr+noCL} & {0.0} & {0.1} & {0.1} & {0.2} & {0.1} & {0.2} & {0.4} & {0.5} \\
{\xlmr+noCL+UMLS$_{\en}$} & {27.6} & {32.0} & {12.2} & {14.7} & {21.8} & {25.9} & {29.3} & {35.9} \\
{\xlmr+noCL+UMLS$_{all}$} & {31.8} & {37.3} & {18.6} & {22.2} & {35.4} & {41.2} & {42.8} & {48.9} \\
{\xlmr+CL} & {14.1} & {17.1} & {5.0} & {6.5} & {8.7} & {11.2} & {21.6} & {27.1} \\
{\xlmr+CL+UMLS$_{\en}$} & {25.2} & {29.0} & {12.1} & {14.1} & {19.8} & {25.0} & {31.1} & {36.1} \\
{\xlmr+CL+UMLS$_{all}$} & {32.1} & {36.7} & {19.1} & {23.8} & {34.9} & {42.4} & {43.4} & {49.0} \\
\hdashline
{\mpnet+noCL} & {19.5} & {25.9} & {12.2} & {14.8} & {19.2} & {24.3} & {28.9} & {36.3} \\
{\mpnet+noCL+UMLS$_{\en}$} & {25.1} & {29.2} & {17.8} & {21.5} & {21.9} & {26.9} & {30.0} & {36.5} \\
{\mpnet+noCL+UMLS$_{all}$} & {\bf 33.4} & {37.8} & {\bf 23.6} & {\bf 27.7} & {\bf 39.8} & {45.4} & {\bf 44.6} & {\bf 51.4} \\
{\mpnet+CL} & {20.8} & {26.5} & {9.1} & {12.5} & {12.8} & {17.1} & {30.4} & {36.5} \\
{\mpnet+CL+UMLS$_{\en}$} & {25.1} & {28.7} & {11.4} & {14.0} & {21.8} & {27.2} & {31.0} & {37.5} \\
{\mpnet+CL+UMLS$_{all}$} & {32.0} & {\bf 38.7} & {22.9} & {27.5} & {39.2} & {\bf 45.7} & {44.3} & {51.0} \\
\bottomrule
\end{tabularx}
}%
\vspace{-1.7mm}
\caption{A summary of results in the XEL task on the biomedical XL-BEL benchmark of \newcite{Liu:2021acl}. We show the results of the better-performing LM (\xlmr), and the more lightweight multilingual SE (\mpnet).}
\label{tab:main-xel}
\vspace{-1.7mm}
\end{table*}
Table~\ref{tab:main-panlex} shows the BLI results on 10 low(er)-resource language pairs from PanLex-BLI. While overall relative trends are similar to those observed for high(er)-resource languages from GT-BLI (Table \ref{tab:main-gtbli}), the gains stemming from cross-lingual contrastive lexical fine-tuning are substantially larger in this case. The best-performing configuration -- contrastive fine-tuning and interpolation (+CL (0.4)) applied on LaBSE -- surpasses \vecmap by 11 BLI points on average (compared to 6 points on GT-BLI), with gains for some language pairs (e.g., \he-\ka, \et-\he) approaching the impressive margin of 20 BLI points. This finding indicates that cross-lingual lexical knowledge stored in multilingual SEs is even more crucial when dealing with lower-resource languages.

In Table~\ref{tab:1k-panlex} we compare the results of LaBSE (as the best-performing multilingual SE) against \vecmap on PanLex-BLI in a scenario with less external bilingual supervision: $|\mathcal{D}| = 1k$. Interestingly, in this setup \labse already substantially outperforms \vecmap out of the box (noCL (1.0)); contrastive lexical fine-tuning (+CL ($1.0$)) and interpolation with \vecmap embeddings (+CL (0.5)) again bring further substantial gains, and we again observe a strong synergistic effect of the two components: +CL (1.0) yields gains over noCL (1.0) for 9/10 language pairs, and +CL (0.5) results in further boosts for all 10 pairs. Further, the contrastively fine-tuned \labse seems to be much more resilient to training data scarcity than \vecmap: reduction of the training dictionary size from 5k to 1k reduces the performance of \labse+CL ($\lambda$) by 27\% (from 34.7 to 25.2 P@1 points) compared to a massive performance drop of almost 70\% for \vecmap (from 23.5 to mere 7.4 P@1).

\rparagraph{Cross-Lingual Lexical Semantic Similarity (XLSIM)}
The average XLSIM results are summarized in Table~\ref{tab:main-gtbli}. They again corroborate one of the main findings from BLI experiments: multilingual SEs store more cross-lingual lexical knowledge than multilingual LMs. This is validated by substantial gains of SEs over corresponding LMs across all configurations in Table~\ref{tab:main-gtbli}. Interestingly, due to their contrastive learning objectives on sentence-level parallel data \cite{Feng:2020labse}, \labse and \mpnet provide very strong XLSIM results when used off-the-shelf (noCL (1.0)), outperforming the CLWE \vecmap embeddings. Contrastive lexical fine-tuning with 5k word translation pairs (+CL (1.0)) in this case does not bring any gains. However, the opposite is true for multilingual LMs: contrastive cross-lingual lexical fine-tuning on only 5k word translation pairs brings large benefits in the XLSIM task (e.g., compare noCL (1.0) and +CL (1.0) configurations for \mbert and \xlmr), and turns them into more effective lexical encoders. This result corroborates a similar finding from prior work in monolingual setups \cite{Vulic:2021acl}. Finally, interpolation with static CLWEs benefits the final XLSIM performance of all four underlying multilingual encoders: interpolated vectors ($\lambda=0.5$) yield highest scores across the board, substantially surpassing gains both \vecmap and WEs from fine-tuned encoders ($\lambda=1.0$).

\rparagraph{Interpolation with Static CLWEs} A more detailed analysis over different $\lambda$ values for BLI and XLSIM, summarized in Figure~\ref{fig:lambdas}, reveals that interpolation can bring large performance gains, especially for $\lambda$ in the $[0.3,0.5]$ interval. The optimal $\lambda$ value is, however, task- and even dataset-dependent. For instance, for low-resource BLI on PanLex-BLI more knowledge comes from the multilingual encoders as \vecmap CLWEs are of lower quality for such languages: in consequence, the optimal $\lambda$ value `moves away' from the static CLWEs towards encodings obtained by fine-tuned multilingual SE. We also note that larger benefits from interpolation are observed when \vecmap CLWEs are combined with contrastively fine-tuned multilingual SEs: cf., the large gains in Figure~\ref{fig:lambda-panlex} and in Table~\ref{tab:1k-panlex} for the \labse+CL model variant.  

\rparagraph{Cross-Lingual Entity Linking (XEL)} Experiments on XL-BEL \citep{Liu:2021acl}, summarized in Table~\ref{tab:main-xel}, demonstrate that additional contrastive tuning with word or phrase pairs can greatly boost performance of multilingual LMs: even fine-tuning with 5k word translation pairs without any domain-specific knowledge yields strong benefits for \xlmr. As expected, using a much larger and domain-specific external database UMLS yields much higher scores and is more crucial for performance. In fact, contrastively fine-tuning on UMLS generally improves XEL performance with all four underlying models. Again, we observe that SE-based (\mpnet) configurations outperform the respective LM-based (\xlmr) configurations across the board. This finding again indicates that multilingual SEs store more cross-lingual lexical knowledge than multilingual LMs: this difference is particularly salient when the models are used off-the-shelf without any additional contrastive fine-tuning, and \mpnet retains the edge over XLM-R even after task-specific fine-tuning with the UMLS-based domain-specific knowledge.

What is more, for \fin, \ru, and \tr, the multilingual \mpnet-based variants match or surpass the performance of respective XEL models trained on top of monolingual LMs (e.g., for \fin, a model based on the Finnish BERT) reported by \citet{Liu:2021acl}. This further validates our hypothesis that multilingual SEs store rich multilingual lexical knowledge, which is then also exposed in domain-specific (multilingual) UMLS fine-tuning, yielding performance gains. Contrastive fine-tuning on UMLS synonyms (+CL+UMLS variants) expectedly outpeforms fine-tuning on (5k) general-domain word translations (+CL), indicating that in specialized domains, if available, in-domain cross-lingual lexical signal should be exploited.

\section{Conclusion and Future Work}
\label{s:conclusion}
In this work, we investigated strategies to expose cross-lingual lexical knowledge from pretrained models, including multilingual language models (LMs) and multilingual sentence encoders (SEs). Based on an extensive evaluation on a suite of cross-lingual lexical tasks, we verified that multilingual SEs (e.g., \labse and \mpnet) are superior to multilingual LMs (\mbert and \xlmr) in terms of stored cross-lingual lexical knowledge. Moreover, we proposed new methods to further fine-tune their representations based on contrastive learning to `rewire' the models' parameters and transform them from LMs and SEs into more effective cross-lingual lexical encoders. We also empirically demonstrated that for some lexical tasks performance can be further boosted through interpolation of type-level lexical encodings with static cross-lingual word embeddings. These yield gains for all underlying models, and are especially significant for resource-poor languages and in low-data learning regimes. While our work has focused on two widely used state-of-the-art multilingual SEs, the contrastive fine-tuning framework is versatile and model-independent and can be directly applied on top of other multilingual SEs in future work. We will also investigate other more sophisticated contrastive learning strategies, look into ensembling of knowledge extracted from different SEs, and expand our evaluation to more tasks and languages.



\section*{Limitations}

This work focuses on lexical specialization of multilingual encoders, off-the-shelf LMs (experiments with mBERT and XLM-R Base) and, in particular multilingual encoders specialized for sentence-level semantics (experiments with \labse and \mpnet). While these are all widely used models, they are arguably among the smaller pretrained multilingual encoders. Due to computational constraints, we have not evaluated the effectiveness of the proposed cross-lingual lexical specialisation for larger multilingual LMs, e.g., XLM-R-Large \cite{Conneau:2020acl} or mT5 (Large, XL, and XXL) \cite{xue2021mt5}. It is possible that these larger multilingual LMs would close (some of) the performance gap w.r.t. multilingual SEs. Such large LMs, however, are effectively available to fewer researchers and practitioners. Our work includes less resource-demanding LMs and SEs, making their lexically specialized variants that we offer, more widely accessible.  

 Lexical input (i.e., words or phrases) are provided to each multilingual encoder fully \textit{``in isolation''} (see \S\ref{s:methodology}), without any surrounding context. However, the alternative of using external corpora and \textit{averaging-over-context} \cite{Litschko:2022jir}, which we have not evaluated in this work for clarity and space constraints, might yield slightly improved task performance. Nonetheless, the 'in isolation' approach has been verified in previous work \cite{Vulic:2021acl,Litschko:2022jir,Li:2022acl} as very competitive, and is more lightweight: \textbf{1)} it disposes of any external text corpora and is not impacted by the external data; \textbf{2)} it encodes words more efficiently due to the absence of context. Moreover, it allows us to directly study the richness of cross-lingual information stored in the encoders' parameters, and its interaction with additional cross-lingual signal from bilingual lexicons.

The contrastive cross-lingual lexical fine-tuning we proposed in this is work is \textit{bilingual}. It leverages a small bilingual dictionary $\mathcal{D}$ for each language pair and specializes the multilingual encoders (LMs and SEs) independently for each language pair. Assuming interest in cross-lingual lexical tasks between all pairs of $N_L$ languages, this entails $\frac{N_L\cdot (N_L - 1)}{2}$ fine-tuning procedures and as many resulting bilingual models. Although our contrastive fine-tuning is relatively fast and lightweight, given that it leverages at most 5k translation pairs, for large $N_L$ it could easily exceed the computational and time budget for most users. 


Intuitively, for each bilingually fine-tuned model, we evaluate the performance for that respective language pair. Currently, we do not investigate the spillover effects that a bilingual lexical fine-tuning of multilingual encoders could have on lexical representations of other languages. Such an analysis, planned for future work, would be particularly interesting in the context of low-resource languages, unseen from the point of view of cross-lingual lexical fine-tuning, and in particular closely related low-resource languages. For instance, if we are doing cross-lingual lexical fine-tuning for language pairs involving Turkish, are there spillover benefits for low(er)-resource Turkic languages such as Uyghur or Kazakh?

Finally, we acknowledge that our choice of lexical tasks as probing tasks is non-exhaustive: we put focus on standard tasks from previous work on (multilingual) lexical semantics that are especially convenient as cross-lingual lexical probes: such tasks directly test and compare the quality of cross-lingual lexical representations obtained via different methods. 

\section*{Acknowledegments}
{\scriptsize\euflag} We would like to thank Yaoyiran Li for interesting insights and discussions. This work has been supported by the ERC PoC Grant MultiConvAI (no. 957356) and a Huawei research
donation to the University of Cambridge. Fangyu Liu is supported by Grace \& Thomas C.H. Chan Cambridge
Scholarship. The work of Goran Glava\v{s} has been supported by the Multi2ConvAI project of MWK Baden-Württemberg. The work of Edoardo Maria Ponti's work has been supported by the Facebook CIFAR AI Chair program during his affiliation with Mila -- Quebec AI Institute. Nigel Collier kindly acknowledges grant-in-aid funding from ESRC (grant number ES/T012277/1).


\bibliography{anthology,custom}
\bibliographystyle{acl_natbib}

\clearpage
\appendix
\section{List of Languages}
\label{app:langs}
The list of languages used in this work, along with their ISO 639-1 codes, is available in Table~\ref{tab:lang_code}.
\begin{table}[!t]
    \centering
    {\footnotesize
    \begin{tabularx}{\columnwidth}{l Y}
    \toprule
    \rowcolor{Gray}
    \multicolumn{2}{l}{{\bf Languages in:} GT-BLI, Multi-SimLex, XL-BEL} \\
         \en & English  \\
         \de & German   \\
         \tr & Turkish  \\
         \fin & Finnish \\
         \hr & Croatian \\
         \ru & Russian \\
         \ita & Italian \\
         \fr & French   \\
         \midrule
         \rowcolor{Gray}
         \multicolumn{2}{l}{{\bf Languages in:} PanLex-BLI} \\
         \blg & Bulgarian \\
         \ca & Catalan \\
         \et &  Estonian \\
         \he & Hebrew \\
         \ka & Georgian \\
    \bottomrule
    \end{tabularx}
    }%
    \caption{Languages and their ISO 639-1 codes.}
  \label{tab:lang_code}
\end{table}

\section{BLI Results across Individual Language Pairs}
\label{app:gtbli}
Additional experiments and analyses over individual language pairs and other $\lambda$ values, which further support the main claims of the paper, have been relegated to the appendix for clarity and compactness of the presentation in the main paper:

\rparagraph{Table~\ref{tab:app-gtbli-nocl}} It provides results over all 28 language pairs in GT-BLI with 2 multilingual LMs and 2 multilingual SEs in the \textit{noCL} variant without contrastive fine-tuning. 

\rparagraph{Table~\ref{tab:app-gtbli-cl}} It provides results over all 28 language pairs in GT-BLI with 2 multilingual LMs and 2 multilingual SEs in the \textit{+CL} variant with contrastive fine-tuning.

\rparagraph{Table~\ref{tab:app-gtbli-lambda-nocl}} It provides results over all 28 language pairs in GT-BLI and across different $\lambda$ values with the \labse+noCL variant. 

\rparagraph{Table~\ref{tab:app-gtbli-lambda-cl}} It provides results over all 28 language pairs in GT-BLI and across different $\lambda$ values with the \labse+CL variant. 

\begin{table*}[t]
\def\arraystretch{0.85}
\centering
{\scriptsize
\begin{tabularx}{\linewidth}{l Y YY YY YY YY}
\toprule
{} & {} & \multicolumn{2}{c}{\bf \mbert} & \multicolumn{2}{c}{\bf \xlmr} & \multicolumn{2}{c}{\bf \labse} & \multicolumn{2}{c}{\bf \mpnet} \\
\cmidrule(lr){3-4} \cmidrule(lr){5-6} \cmidrule(lr){7-8} \cmidrule(lr){9-10}
\rowcolor{Gray}
{\bf Pair} $\downarrow$ {\bf / Config} $\xrightarrow{}$ & {\bf \vecmap} & {+noCL (1.0)} & {+noCL (0.3)} & {+noCL (1.0)} & {+noCL (0.3)} & {+noCL (1.0)} & {+noCL (0.3)} & {+noCL (1.0)} & {+noCL (0.3)} \\
\cmidrule(lr){3-4} \cmidrule(lr){5-6} \cmidrule(lr){7-8} \cmidrule(lr){9-10}
 \en--\de & {55.6} & {15.6} & {50.7} & {12.7} & {45.5} & {25.4} & {54.1} & {22.0} & {47.7} \\ 
 \en--\tr & {40.4} & {7.2} & {34.9} & {6.0} & {28.9} & {23.6} & {42.1} & {16.1} & {33.7} \\ 
 \en--\fin & {45.6} & {7.9} & {38.7} & {6.6} & {33.4} & {19.3} & {45.1} & {14.8} & {39.5} \\ 
 \en--\hr & {37.5} & {8.9} & {31.8} & {6.8} & {25.6} & {24.7} & {45.3} & {18.5} & {38.9} \\ 
 \en--\ru & {45.6} & {3.2} & {40.7} & {0.9} & {34.7} & {24.7} & {49.9} & {17.6} & {41.7} \\ 
 \en--\ita & {60.2} & {12.3} & {57.1} & {9.3} & {53.0} & {26.4} & {62.3} & {23.5} & {58.8} \\ 
 \en--\fr & {64.1} & {25.2} & {62.5} & {19.5} & {56.6} & {34.2} & {67.5} & {29.2} & {61.7} \\ 
 \de--\tr & {32.5} & {9.1} & {28.9} & {6.9} & {24.4} & {17.7} & {33.0} & {13.1} & {29.7} \\ 
 \de--\fin & {39.7} & {9.2} & {34.1} & {7.3} & {30.1} & {16.0} & {37.4} & {13.3} & {34.7} \\ 
 \de--\hr & {33.3} & {11.5} & {31.0} & {9.7} & {25.7} & {19.2} & {38.5} & {14.9} & {34.1} \\ 
 \de--\ru & {40.0} & {4.2} & {36.4} & {0.9} & {32.4} & {14.3} & {41.5} & {9.5} & {37.7} \\ 
 \de--\ita & {49.5} & {10.9} & {45.9} & {8.1} & {42.7} & {19.4} & {51.4} & {18.3} & {49.1} \\ 
 \de--\fr & {50.0} & {15.8} & {49.7} & {10.5} & {42.3} & {22.5} & {53.2} & {20.2} & {49.8} \\ 
 \tr--\fin & {31.3} & {6.7} & {26.2} & {5.2} & {22.0} & {15.5} & {31.7} & {11.3} & {30.9} \\ 
 \tr--\hr & {25.4} & {10.8} & {24.3} & {8.3} & {20.5} & {18.7} & {33.1} & {14.4} & {28.2} \\ 
 \tr--\ru & {32.9} & {2.6} & {29.1} & {0.8} & {25.5} & {14.1} & {36.9} & {11.3} & {33.5} \\ 
 \tr--\ita & {37.1} & {7.9} & {34.7} & {5.5} & {27.4} & {17.0} & {38.9} & {14.8} & {38.3} \\ 
 \tr--\fr & {39.4} & {7.8} & {37.3} & {5.7} & {30.9} & {20.9} & {43.1} & {16.6} & {39.4} \\ 
 \fin--\hr & {30.4} & {7.2} & {26.6} & {5.5} & {22.8} & {17.4} & {36.4} & {12.2} & {32.7} \\ 
 \fin--\ru & {38.2} & {2.6} & {34.0} & {0.9} & {30.5} & {15.1} & {41.0} & {9.0} & {37.1} \\ 
 \fin--\ita & {39.9} & {7.9} & {36.7} & {6.8} & {30.4} & {18.1} & {43.4} & {16.4} & {41.8} \\ 
 \fin--\fr & {42.8} & {7.5} & {38.9} & {5.9} & {32.2} & {18.6} & {45.9} & {16.2} & {42.2} \\ 
  \hr--\ru & {40.6} & {6.0} & {35.8} & {1.6} & {30.4} & {24.5} & {45.7} & {16.3} & {41.4} \\ 
 \hr--\ita & {40.4} & {11.2} & {39.0} & {8.4} & {31.3} & {24.5} & {47.9} & {22.4} & {44.5} \\ 
 \hr--\fr & {43.6} & {9.7} & {42.3} & {6.1} & {30.6} & {25.7} & {50.0} & {19.8} & {43.9} \\ 
 \ru--\ita & {46.6} & {3.1} & {42.2} & {1.5} & {36.4} & {21.8} & {47.6} & {18.4} & {46.5} \\ 
 \ru--\fr & {48.7} & {4.1} & {44.3} & {1.5} & {38.4} & {26.6} & {50.9} & {18.9} & {47.5} \\ 
 \ita--\fr & {64.1} & {16.6} & {62.8} & {9.3} & {58.6} & {33.9} & {65.6} & {27.5} & {63.1} \\
 \midrule
 {\bf Average} & {42.7} & {9.0} & {39.2} & {6.4} & {33.7} & {21.4} & {45.7} & {17.0} & {41.7} \\
\bottomrule
\end{tabularx}
}%
\caption{Individual P@1 scores ($\times$100\%) for all 28 language pairs in the GT-BLI dataset of \newcite{Glavas:2019acl}, with multilingual LMs and SEs used `off-the-shelf' \textit{without contrastive fine-tuning} (\S\ref{s:methodology}). See \S\ref{s:exp} for the description of different model configurations/variants. $|\mathcal{D}|=5k$. The number in the parentheses denotes the value for $\lambda$ (see \S\ref{s:exp}): the value of 1.0 effectively means 'no interpolation' with static \vecmap CLWEs.}
\label{tab:app-gtbli-nocl}
\end{table*}

\begin{table*}[!t]
\def\arraystretch{0.85}
\centering
{\scriptsize
\begin{tabularx}{\linewidth}{l Y YY YY YY YY}
\toprule
{} & {} & \multicolumn{2}{c}{\bf \mbert} & \multicolumn{2}{c}{\bf \xlmr} & \multicolumn{2}{c}{\bf \labse} & \multicolumn{2}{c}{\bf \mpnet} \\
\cmidrule(lr){3-4} \cmidrule(lr){5-6} \cmidrule(lr){7-8} \cmidrule(lr){9-10}
\rowcolor{Gray}
{\bf Pair} $\downarrow$ {\bf / Config} $\xrightarrow{}$ & {\bf \vecmap} & {+CL (1.0)} & {+CL (0.3)} & {+CL (1.0)} & {+CL (0.3)} & {+CL (1.0)} & {+CL (0.3)} & {+CL (1.0)} & {+CL (0.3)} \\
\cmidrule(lr){3-4} \cmidrule(lr){5-6} \cmidrule(lr){7-8} \cmidrule(lr){9-10}
 \en--\de & {55.6} & {26.4} & {59.2} & {24.5} & {56.3} & {31.6} & {61.1} & {30.2} & {58.7} \\ 
 \en--\tr & {40.4} & {17.4} & {39.3} & {19.2} & {42.0} & {33.1} & {50.1} & {30.4} & {45.7} \\ 
 \en--\fin & {45.6} & {18.6} & {45.6} & {20.2} & {45.7} & {30.8} & {53.3} & {28.7} & {50.4} \\ 
 \en--\hr & {37.5} & {23.6} & {44.3} & {21.8} & {44.3} & {36.9} & {53.9} & {31.8} & {49.6} \\ 
 \en--\ru & {45.6} & {23.9} & {48.9} & {28.3} & {48.8} & {46.4} & {55.8} & {37.9} & {53.1} \\ 
 \en--\ita & {60.2} & {26.8} & {64.0} & {25.4} & {61.7} & {33.3} & {66.9} & {33.0} & {65.3} \\ 
 \en--\fr & {64.1} & {34.4} & {67.7} & {33.0} & {65.4} & {42.1} & {71.2} & {39.5} & {68.5} \\ 
 \de--\tr & {32.5} & {19.5} & {32.8} & {15.5} & {33.5} & {24.4} & {37.6} & {22.7} & {36.2} \\ 
 \de--\fin & {39.7} & {20.3} & {38.5} & {19.0} & {37.5} & {25.0} & {43.3} & {24.4} & {42.1} \\ 
 \de--\hr & {33.3} & {24.2} & {37.3} & {22.4} & {36.9} & {27.9} & {42.9} & {27.2} & {41.8} \\ 
 \de--\ru & {40.0} & {21.2} & {42.0} & {19.1} & {41.5} & {27.5} & {45.6} & {26.8} & {44.1} \\ 
 \de--\ita & {49.5} & {23.0} & {49.9} & {19.0} & {48.6} & {24.6} & {52.5} & {25.6} & {51.4} \\ 
 \de--\fr & {50.0} & {27.4} & {52.5} & {24.3} & {51.5} & {31.3} & {54.5} & {30.0} & {53.9} \\ 
 \tr--\fin & {31.3} & {18.0} & {29.6} & {15.9} & {31.7} & {22.2} & {34.9} & {22.4} & {35.9} \\ 
 \tr--\hr & {25.4} & {21.8} & {30.2} & {19.0} & {30.8} & {27.4} & {36.8} & {26.0} & {36.4} \\ 
 \tr--\ru & {32.9} & {16.2} & {33.9} & {15.7} & {33.5} & {24.7} & {37.4} & {24.7} & {37.8} \\ 
 \tr--\ita & {37.1} & {17.4} & {37.8} & {15.4} & {36.8} & {22.8} & {41.6} & {22.2} & {41.0} \\ 
 \tr--\fr & {39.4} & {18.4} & {40.4} & {17.4} & {38.9} & {29.4} & {43.9} & {26.3} & {43.6} \\ 
 \fin--\hr & {30.4} & {20.2} & {32.7} & {17.4} & {32.5} & {27.4} & {39.7} & {24.3} & {38.4} \\ 
 \fin--\ru & {38.2} & {17.6} & {37.3} & {18.0} & {38.7} & {27.6} & {42.4} & {27.6} & {41.3} \\ 
 \fin--\ita & {39.9} & {18.4} & {40.5} & {17.6} & {40.0} & {25.2} & {45.3} & {24.0} & {44.9} \\ 
 \fin--\fr & {42.8} & {17.5} & {43.0} & {18.2} & {41.9} & {26.4} & {47.6} & {26.1} & {45.8} \\ 
 \hr--\ru & {40.6} & {26.4} & {41.4} & {29.5} & {43.9} & {36.1} & {46.4} & {33.3} & {46.4} \\ 
 \hr--\ita & {40.4} & {24.6} & {44.0} & {22.8} & {42.8} & {32.3} & {49.7} & {30.2} & {49.2} \\ 
 \hr--\fr & {43.6} & {24.1} & {46.5} & {23.4} & {44.1} & {35.0} & {51.8} & {29.3} & {50.7} \\ 
 \ru--\ita & {46.6} & {22.0} & {46.9} & {19.4} & {45.0} & {32.0} & {50.2} & {28.4} & {51.1} \\ 
 \ru--\fr & {48.7} & {22.1} & {49.1} & {20.5} & {47.6} & {37.0} & {53.5} & {28.9} & {51.2} \\ 
 \ita--\fr & {64.1} & {33.4} & {65.5} & {32.9} & {64.3} & {42.2} & {66.0} & {39.0} & {66.4} \\ 
 \midrule
 {\bf Average} & {42.7} & {22.3} & {44.3} & {21.2} & {43.8} & {30.8} & {49.1} & {28.6} & {47.9} \\ 
\bottomrule
\end{tabularx}
}%
\caption{Individual P@1 scores ($\times$100\%) for all 28 language pairs in the GT-BLI dataset of \newcite{Glavas:2019acl}, with model variants \textit{with contrastive fine-tuning} (\S\ref{s:methodology}). $|\mathcal{D}|=5k$.}
\label{tab:app-gtbli-cl}
\end{table*}

\begin{table*}[!t]
\def\arraystretch{0.85}
\centering
{\scriptsize
\begin{tabularx}{\linewidth}{l Y YYYYYYY}
\toprule
{} & {} & \multicolumn{7}{c}{\bf \labse+noCL} \\
\cmidrule(lr){3-9}
\rowcolor{Gray}
{\bf Pair} $\downarrow$ {\bf / $\lambda=$} $\xrightarrow{}$ & {\bf \vecmap}(0.0) & {0.1} & {0.2} & {0.3} & {0.4} & {0.5} & {0.7} & {1.0} \\
\cmidrule(lr){3-9}
\en--\de & {55.6} & {57.3} & {56.3} & {54.1} & {50.9} & {47.8} & {40.6} & {25.4} \\ 
\en--\tr & {40.4} & {42.0} & {41.9} & {42.1} & {41.6} & {40.2} & {35.9} & {23.6} \\ 
\en--\fin & {45.6} & {46.1} & {46.1} & {45.1} & {44.4} & {42.8} & {34.4} & {19.3} \\ 
\en--\hr & {37.5} & {39.8} & {43.0} & {45.3} & {46.5} & {46.5} & {41.3} & {24.7} \\ 
\en--\ru & {45.6} & {46.5} & {48.0} & {49.9} & {50.5} & {50.2} & {46.1} & {24.7} \\ 
\en--\ita & {60.2} & {61.6} & {63.0} & {62.3} & {61.3} & {60.1} & {49.9} & {26.4} \\ 
\en--\fr & {64.1} & {66.0} & {67.1} & {67.5} & {65.9} & {64.8} & {55.3} & {34.2} \\ 
\de--\tr & {32.5} & {33.6} & {33.4} & {33.0} & {32.3} & {30.2} & {24.4} & {17.7} \\ 
\de--\fin & {39.7} & {39.7} & {39.3} & {37.4} & {35.9} & {34.3} & {25.8} & {16.0} \\ 
\de--\hr & {33.3} & {35.8} & {37.0} & {38.5} & {38.8} & {36.1} & {29.2} & {19.2} \\ 
\de--\ru & {40.0} & {40.9} & {41.2} & {41.5} & {41.1} & {38.4} & {29.6} & {14.3} \\ 
\de--\ita & {49.5} & {51.3} & {51.3} & {51.4} & {49.4} & {46.3} & {36.4} & {19.4} \\ 
\de--\fr & {50.0} & {52.2} & {53.2} & {53.2} & {51.8} & {49.1} & {37.8} & {22.5} \\ 
\tr--\fin & {31.3} & {32.3} & {31.6} & {31.7} & {31.2} & {29.8} & {24.8} & {15.5} \\ 
\tr--\hr & {25.4} & {28.3} & {31.2} & {33.1} & {34.1} & {33.6} & {28.9} & {18.7} \\ 
\tr--\ru & {32.9} & {34.9} & {35.5} & {36.9} & {36.5} & {34.0} & {29.0} & {14.1} \\ 
\tr--\ita & {37.1} & {38.7} & {39.6} & {38.9} & {38.8} & {37.7} & {31.5} & {17.0} \\ 
\tr--\fr & {39.4} & {41.4} & {42.1} & {43.1} & {43.4} & {41.9} & {34.6} & {20.9} \\ 
\fin--\hr & {30.4} & {32.3} & {34.7} & {36.4} & {36.9} & {36.3} & {28.6} & {17.4} \\ 
\fin--\ru & {38.2} & {39.5} & {40.0} & {41.0} & {40.5} & {37.6} & {28.9} & {15.1} \\ 
\fin--\ita & {39.9} & {42.9} & {42.9} & {43.4} & {44.2} & {41.5} & {34.0} & {18.1} \\ 
\fin--\fr & {42.8} & {44.7} & {45.9} & {45.9} & {46.1} & {43.6} & {34.6} & {18.6} \\ 
\hr--\ru & {40.6} & {41.8} & {43.9} & {45.7} & {45.8} & {45.0} & {39.0} & {24.5} \\ 
\hr--\ita & {40.4} & {43.5} & {46.0} & {47.9} & {48.6} & {47.6} & {41.8} & {24.5} \\ 
\hr--\fr & {43.6} & {46.8} & {48.6} & {50.0} & {50.1} & {47.9} & {42.0} & {25.7} \\ 
\ru--\ita  & {46.6} & {48.1} & {48.1} & {47.6} & {46.8} & {44.5} & {38.7} & {21.8} \\ 
\ru--\fr & {48.7} & {50.2} & {51.0} & {50.9} & {50.0} & {49.0} & {43.6} & {26.6} \\ 
\ita--\fr & {64.1} & {64.9} & {65.8} & {65.6} & {64.9} & {63.0} & {54.0} & {33.9} \\ 
\midrule
 {\bf Average} & {42.7} & {44.4} & {45.3} & {45.7} & {45.3} & {43.6} & {36.5} & {21.4} \\ 
\bottomrule
\end{tabularx}
}%
\caption{Individual P@1 scores ($\times$100\%) for all 28 language pairs in the GT-BLI dataset of \newcite{Glavas:2019acl}, across different values for $\lambda$. The model variant is \labse+noCL (see \S\ref{s:exp}); similar patterns are observed with another multilingual SE in our evaluation (\mpnet). $|\mathcal{D}|=5k$.}
\label{tab:app-gtbli-lambda-nocl}
\end{table*}

\begin{table*}[!t]
\def\arraystretch{0.85}
\centering
{\scriptsize
\begin{tabularx}{\linewidth}{l Y YYYYYYY}
\toprule
{} & {} & \multicolumn{7}{c}{\bf \labse+CL} \\
\cmidrule(lr){3-9}
\rowcolor{Gray}
{\bf Pair} $\downarrow$ {\bf / $\lambda=$} $\xrightarrow{}$ & {\bf \vecmap}(0.0) & {0.1} & {0.2} & {0.3} & {0.4} & {0.5} & {0.7} & {1.0} \\
\cmidrule(lr){3-9}
 \en--\de & {55.6} & {59.4} & {61.2} & {61.1} & {60.1} & {56.8} & {46.5} & {31.6} \\ 
 \en--\tr & {40.4} & {43.9} & {46.8} & {50.1} & {51.0} & {49.9} & {45.4} & {33.1} \\ 
\en--\fin & {45.6} & {48.6} & {51.0} & {53.3} & {54.1} & {53.9} & {46.1} & {30.8} \\ 
\en--\hr & {37.5} & {44.1} & {49.7} & {53.9} & {56.7} & {57.1} & {49.7} & {36.9} \\ 
\en--\ru & {45.6} & {50.1} & {52.5} & {55.8} & {58.5} & {59.0} & {56.2} & {46.4} \\ 
 \en--\ita & {60.2} & {62.5} & {65.3} & {66.9} & {67.2} & {66.2} & {55.7} & {33.3} \\ 
 \en--\fr & {64.1} & {66.3} & {69.5} & {71.2} & {71.1} & {69.5} & {59.5} & {42.1} \\ 
 \de--\tr & {32.5} & {35.4} & {36.5} & {37.6} & {36.7} & {35.6} & {31.8} & {24.4} \\ 
\de--\fin & {39.7} & {41.6} & {42.5} & {43.3} & {42.2} & {39.5} & {32.7} & {25.0} \\ 
\de--\hr & {33.3} & {37.9} & {40.8} & {42.9} & {43.1} & {41.6} & {35.9} & {27.9} \\ 
\de--\ru & {40.0} & {43.1} & {44.0} & {45.6} & {45.9} & {44.7} & {37.3} & {27.5} \\ 
\de--\ita & {49.5} & {51.3} & {52.0} & {52.5} & {50.9} & {47.7} & {38.9} & {24.6} \\ 
\de--\fr & {50.0} & {52.4} & {53.7} & {54.5} & {53.6} & {50.3} & {42.7} & {31.3} \\ 
\tr--\fin & {31.3} & {33.3} & {34.3} & {34.9} & {35.0} & {33.7} & {30.2} & {22.2} \\ 
\tr--\hr & {25.4} & {30.4} & {34.2} & {36.8} & {38.5} & {38.2} & {35.8} & {27.4} \\ 
\tr--\ru & {32.9} & {35.2} & {36.6} & {37.4} & {37.5} & {36.1} & {32.5} & {24.7} \\ 
\tr--\ita & {37.1} & {39.0} & {41.2} & {41.6} & {41.5} & {39.9} & {34.4} & {22.8} \\ 
\tr--\fr & {39.4} & {41.8} & {43.0} & {43.9} & {43.8} & {43.3} & {38.6} & {29.4} \\ 
\fin--\hr & {30.4} & {34.0} & {37.5} & {39.7} & {41.2} & {40.2} & {36.4} & {27.4} \\ 
\fin--\ru & {38.2} & {40.2} & {41.1} & {42.4} & {42.6} & {40.2} & {36.2} & {27.6} \\ 
\fin--\ita & {39.9} & {43.3} & {44.3} & {45.3} & {45.9} & {44.7} & {38.2} & {25.2} \\ 
\fin--\fr & {42.8} & {44.5} & {46.3} & {47.6} & {47.8} & {46.0} & {40.0} & {26.4} \\ 
\hr--\ru & {40.6} & {42.0} & {44.9} & {46.4} & {47.5} & {48.3} & {44.7} & {36.1} \\ 
\hr--\ita & {40.4} & {43.3} & {47.7} & {49.7} & {50.6} & {50.4} & {46.1} & {32.3} \\ 
\hr--\fr & {43.6} & {47.2} & {49.0} & {51.8} & {52.0} & {51.0} & {46.4} & {35.0} \\ 
\ru--\ita & {46.6} & {48.8} & {49.6} & {50.2} & {50.4} & {48.7} & {44.9} & {32.0} \\ 
\ru--\fr & {48.7} & {51.0} & {51.9} & {53.5} & {53.1} & {52.1} & {47.1} & {37.0} \\ 
\ita--\fr & {64.1} & {64.9} & {65.9} & {66.0} & {66.0} & {64.6} & {56.9} & {42.2} \\ 
\midrule
{\bf Average} & {42.7} & {45.6} & {47.6} & {49.1} & {49.4} & {48.2} & {42.4} & {30.8} \\
\bottomrule
\end{tabularx}
}%
\caption{Individual P@1 scores ($\times$100\%) for all 28 language pairs in the GT-BLI dataset of \newcite{Glavas:2019acl}, across different values for $\lambda$. The model variant is \labse+CL (see \S\ref{s:exp}); similar patterns are observed with another multilingual SE in our evaluation (\mpnet). $|\mathcal{D}|=5k$.}
\label{tab:app-gtbli-lambda-cl}
\end{table*}


\section{XLSIM Results across Individual Language Pairs}
\label{app:msimlex}
\sparagraph{Table~\ref{tab:app-msimlex-nocl}} It provides results over selected 6 language pairs from Multi-SimLex with 2 multilingual LMs and 2 multilingual SEs in the \textit{noCL} variant without contrastive fine-tuning. 

\rparagraph{Table~\ref{tab:app-msimlex-cl}} It provides results over selected 6 language pairs from Multi-SimLex with 2 multilingual LMs and 2 multilingual SEs in the \textit{+CL} variant with contrastive fine-tuning.

\rparagraph{Table~\ref{tab:app-msimlex-lambda-nocl}} It provides results over selected 6 language pairs from Multi-SimLex and across different $\lambda$ values with the \mpnet+noCL variant. 

\rparagraph{Table~\ref{tab:app-msimlex-lambda-cl}} It provides results over selected 6 language pairs from Multi-SimLex and across different $\lambda$ values with the \mpnet+CL variant.

\begin{table*}[t]
\def\arraystretch{0.85}
\centering
{\scriptsize
\begin{tabularx}{\linewidth}{l Y YY YY YY YY}
\toprule
{} & {} & \multicolumn{2}{c}{\bf \mbert} & \multicolumn{2}{c}{\bf \xlmr} & \multicolumn{2}{c}{\bf \labse} & \multicolumn{2}{c}{\bf \mpnet} \\
\cmidrule(lr){3-4} \cmidrule(lr){5-6} \cmidrule(lr){7-8} \cmidrule(lr){9-10}
\rowcolor{Gray}
{\bf Pair} $\downarrow$ {\bf / Config} $\xrightarrow{}$ & {\bf \vecmap} & {+noCL (1.0)} & {+noCL (0.5)} & {+noCL (1.0)} & {+noCL (0.5)} & {+noCL (1.0)} & {+noCL (0.5)} & {+noCL (1.0)} & {+noCL (0.5)} \\
\cmidrule(lr){3-4} \cmidrule(lr){5-6} \cmidrule(lr){7-8} \cmidrule(lr){9-10}
 \en--\fin & 42.2 & {1.1} & {30.9} & {1.4} & {21.4} & {46.8} & {51.6} & {47.7} & {53.4} \\ 
 \en--\ru & 39.7 & {5.2} & {30.1} & {2.5} & {21.3} & {53.5} & {52.3} & {57.5} & {55.2} \\ 
 \en--\fr & 64.8 & {11.3} & {52.8} & {2.2} & {29.6} & {68.5} & {74.2} & {64.7} & {73.9} \\ 
 \fin--\ru & 33.0 & {4.3} & {25.3} & {3.2} & {20.7} & {38.4} & {41.6} & {42.7} & {45.6} \\ 
 \fin--\fr & 46.7 & {3.6} & {35.3} & {0.3} & {22.8} & {43.1} & {52.6} & {42.9} & {53.2} \\ 
 \ru--\fr & 48.2 & {8.6} & {37.9} & {0.8} & {25.3} & {52.2} & {57} & {52.3} & {58.4} \\ 
 \midrule
 {\bf Average} & 45.8 & {5.7} & {35.4} & {1.7} & {23.5} & {50.4} & {54.9} & {51.3} & {56.6} \\
\bottomrule
\end{tabularx}
}%
\caption{Individual Spearman's $\rho$ correlation scores ($\times$100) on the XLSIM task (Multi-SimLex) for a subset of language pairs in our evaluation, with multilingual LMs and SEs used `off-the-shelf' \textit{without contrastive fine-tuning} (\S\ref{s:methodology}). See \S\ref{s:exp} for the description of different model configurations/variants. $|\mathcal{D}|=5k$, with XLSIM test pairs removed from the dictionary. The number in the parentheses denotes the value for $\lambda$ (see \S\ref{s:exp}).}
\label{tab:app-msimlex-nocl}
\end{table*}

\begin{table*}[!t]
\def\arraystretch{0.85}
\centering
{\scriptsize
\begin{tabularx}{\linewidth}{l Y YY YY YY YY}
\toprule
{} & {} & \multicolumn{2}{c}{\bf \mbert} & \multicolumn{2}{c}{\bf \xlmr} & \multicolumn{2}{c}{\bf \labse} & \multicolumn{2}{c}{\bf \mpnet} \\
\cmidrule(lr){3-4} \cmidrule(lr){5-6} \cmidrule(lr){7-8} \cmidrule(lr){9-10}
\rowcolor{Gray}
{\bf Pair} $\downarrow$ {\bf / Config} $\xrightarrow{}$ & {\bf \vecmap} & {+noCL (1.0)} & {+noCL (0.5)} & {+noCL (1.0)} & {+noCL (0.5)} & {+noCL (1.0)} & {+noCL (0.5)} & {+noCL (1.0)} & {+noCL (0.5)} \\
\cmidrule(lr){3-4} \cmidrule(lr){5-6} \cmidrule(lr){7-8} \cmidrule(lr){9-10}
 \en--\fin & {42.2} & {32.4} & {43.5} & {42.3} & {48} & {45.6} & {50.6} & {48.1} & {51.5} \\ 
 \en--\ru & {39.7} & {34.8} & {42.1} & {45.8} & {47.3} & {47.1} & {49.2} & {50.5} & {50.4} \\ 
\en--\fr & {64.8} & {56.5} & {67.4} & {57.9} & {69.2} & {64} & {72} & {64.1} & {71.9} \\ 
 \fin--\ru & {33.0} & {28.4} & {35.9} & {38.3} & {40.8} & {38.3} & {42.3} & {40.1} & {43.1} \\ 
 \fin--\fr & {46.7} & {34.7} & {47.3} & {41.7} & {50.6} & {45.9} & {53.5} & {46.6} & {54.2} \\ 
 \ru--\fr & {48.2} & {43.6} & {52.1} & {50.4} & {55.1} & {51.8} & {56.8} & {48.5} & {55.6} \\ 
 \midrule
 {\bf Average} & {45.8} & {38.4} & {48.1} & {46.1} & {51.8} & {48.8} & {54.1} & {49.6} & {54.5} \\ 
\bottomrule
\end{tabularx}
}%
\caption{Individual Spearman's $\rho$ correlation scores ($\times$100) on the XLSIM task (Multi-SimLex) for a subset of language pairs in our evaluation, with model variants \textit{with contrastive fine-tuning} (\S\ref{s:methodology}). $|\mathcal{D}|=5k$.}
\label{tab:app-msimlex-cl}
\end{table*}

\begin{table*}[!t]
\def\arraystretch{0.85}
\centering
{\scriptsize
\begin{tabularx}{\linewidth}{l Y YYYYYYYY}
\toprule
{} & {} & \multicolumn{8}{c}{\bf \mpnet+noCL} \\
\cmidrule(lr){2-10}
\rowcolor{Gray}
{\bf Pair} $\downarrow$ {\bf / $\lambda=$} $\xrightarrow{}$ & {0.0} & {0.1} & {0.2} & {0.3} & {0.4} & {0.5} & {0.6} & {0.7} & {1.0} \\
\cmidrule(lr){2-10}
  \en--\fin & {42.2} & {45.2} & {48.2} & {50.7} & {52.6} & {53.4} & {53} & {52.2} & {47.7} \\ 
 \en--\ru & {39.7} & {43.1} & {46.6} & {49.9} & {52.9} & {55.2} & {56.7} & {57.5} & {57.5} \\ 
 \en--\fr & {64.8} & {67.6} & {70.1} & {72.2} & {73.5} & {73.9} & {73.2} & {71.7} & {64.7} \\ 
 \fin--\ru & {33} & {36.1} & {39.3} & {42.5} & {44.7} & {45.6} & {45.5} & {45.4} & {42.7} \\ 
\fin--\fr & {46.7} & {49.2} & {51.5} & {53.1} & {53.6} & {53.2} & {51.5} & {49.6} & {42.9} \\ 
 \ru--\fr & {48.2} & {51.1} & {53.8} & {56.2} & {57.7} & {58.4} & {58.1} & {57.2} & {52.3} \\ 
 \midrule
 {\bf Average} & {45.8} & {48.7} & {51.6} & {54.1} & {55.8} & {56.6} & {56.3} & {55.6} & {51.3} \\ 
\bottomrule
\end{tabularx}
}%
\caption{Individual Spearman's $\rho$ correlation scores ($\times$100) on the XLSIM task (Multi-SimLex) for a subset of language pairs in our evaluation, across different values for $\lambda$. The model variant is \mpnet+noCL (see \S\ref{s:exp}); similar patterns are observed with another multilingual SE in our evaluation (\labse). $|\mathcal{D}|=5k$.}
\label{tab:app-msimlex-lambda-nocl}
\end{table*}

\begin{table*}[!t]
\def\arraystretch{0.85}
\centering
{\scriptsize
\begin{tabularx}{\linewidth}{l Y YYYYYYYY}
\toprule
{} & {} & \multicolumn{8}{c}{\bf \mpnet+CL} \\
\cmidrule(lr){2-10}
\rowcolor{Gray}
{\bf Pair} $\downarrow$ {\bf / $\lambda=$} $\xrightarrow{}$ & {0.0} & {0.1} & {0.2} & {0.3} & {0.4} & {0.5} & {0.6} & {0.7} & {1.0} \\
\cmidrule(lr){2-10}
\en--\fin & {42.2} & {44.2} & {46.4} & {48.6} & {50.4} & {51.5} & {51.8} & {51.3} & {48.1} \\ 
\en--\ru & {39.7} & {41.6} & {43.9} & {46.3} & {48.6} & {50.4} & {51.6} & {52} & {50.5} \\ 
\en--\fr & {64.8} & {66.9} & {69} & {70.7} & {71.8} & {71.9} & {71} & {69.5} & {64.1} \\ 
\fin--\ru & {33.0} & {35.1} & {37.5} & {39.9} & {41.9} & {43.1} & {43.5} & {43} & {40.1} \\ 
\fin--\fr & {46.7} & {48.9} & {51.2} & {53.1} & {54.2} & {54.2} & {53.3} & {51.7} & {46.6} \\ 
 \ru--\fr & {48.2} & {50.1} & {52.1} & {53.8} & {55.2} & {55.6} & {55.1} & {53.8} & {48.5} \\ 
 \midrule
 {\bf Average} & {45.8} & {47.8} & {50} & {52.1} & {53.7} & {54.5} & {54.4} & {53.6} & {49.6} \\ 
\bottomrule
\end{tabularx}
}%
\caption{Individual Spearman's $\rho$ correlation scores ($\times$100) on the XLSIM task (Multi-SimLex) for a subset of language pairs in our evaluation, across different values for $\lambda$. The model variant is \mpnet+CL (see \S\ref{s:exp}); similar patterns are observed with another multilingual SE in our evaluation (\labse). $|\mathcal{D}|=5k$.}
\label{tab:app-msimlex-lambda-cl}
\end{table*}

\section{Models and Evaluation Data}
URLs to the models used in this paper are provided in Table~\ref{tab:models}. Training and test data for all three tasks (BLI, XLSIM, XEL) is available online:
\begin{itemize}
    \item GT-BLI is available here: {\small \url{https://github.com/codogogo/xling-eval}}
    \item PanLex-BLI: {\small \url{https://github.com/cambridgeltl/panlex-bli}}
    \item Multi-SimLex [XLSIM]: {\small \url{https://multisimlex.com/}}
    \item XL-BEL [XEL]: {\small \url{https://github.com/cambridgeltl/sapbert}}
\end{itemize}

Our code is based on PyTorch, and relies on the following two widely used repositories:
\begin{itemize*}
    \item {\small \texttt{sentence-transformers:}} {\small \url{www.sbert.net}}
    \item {\small \url{huggingface.co/transformers/}}
\end{itemize*}

\begin{table*}[!t]
\def\arraystretch{0.99}
\centering
{\footnotesize
\begin{tabularx}{\textwidth}{l X}
\toprule
{\bf Name} & {\bf URL} \\
\cmidrule(lr){1-1} \cmidrule(lr){2-2}
{\mbert} & {\small \url{huggingface.co/bert-base-multilingual-uncased}} \\
{\xlmr} & {\small \url{huggingface.co/xlm-roberta-base}} \\
{\labse} & {\small \url{huggingface.co/sentence-transformers/LaBSE}} \\
{\mpnet} & {\small \url{huggingface.co/sentence-transformers/paraphrase-multilingual-mpnet-base-v2}} \\
\bottomrule
\end{tabularx}
}
\vspace{-1.5mm}
\caption{URLs of the multilingual Transformer models used in this work.}
\label{tab:models}
\end{table*}

\end{document}